%% file: manuscript.tex
\documentclass[12pt]{article}

\usepackage{preamble}
\usepackage{variables}

\begin{document}

\pagenumbering{gobble}

\renewcommand*{\thefootnote}{\fnsymbol{footnote}}

\clearpage
\begin{titlepage}
    \begin{center}
        
        \Large
        \textbf{\papertitle}
        
        \vspace{2cm}
        
        \normalsize
        Nicholas J. Hallman\\
        University of Texas at Austin\\
        NicholasHallman@utexas.edu\\
        
        \vspace{.5cm}
        
        Zachary T. Kowaleski\footnote{Corresponding author. University of Texas at Austin. 2110 Speedway, Austin, Texas 78705.}\\
        University of Texas at Austin\\
        zkowaleski@utexas.edu\\
        
        \vspace{.5cm}

        Anu Puvvada\\
        KPMG LLP\\
        Anu.Puvvada@kpmg.com\\
        
        \vspace{.5cm}

        Jaime J. Schmidt\\
        University of Texas at Austin\\
        Jaime.Schmidt@mccombs.utexas.edu\\
        
        \vspace{1.0cm}
        
        \mmddyyyydate
        Draft Date: \formatdate{25}{8}{2026}
        
        \vspace{1.0cm}

        \vfill
    
    \end{center}
    \small
    We thank Kevin Avalos, Ethan Burris, Steve Chase, Tianhui Gou, David Harrison, Steve Kachelmeier, Clay Kohler, Xinxuan Li (discussant), Kevin Veenstra (discussant), Adam Weiss, and Olivia Weiss for guidance, feedback, and data support. We thank participants at the Annual Telfer Conference in Accounting, Auditing, and Accountability and the Arizona Accounting Research Conference for helpful comments and suggestions. Jaime Schmidt gratefully acknowledges financial support from the KPMG Centennial Fellowship in Accounting. Anu Puvvada is a full time employee of KPMG LLP. The remaining authors declare that they have no conflicts of interest. The proprietary data used in this paper were provided by KPMG LLP and classified using a large language model, as described in the paper. We are grateful to KPMG LLP for providing data access.

    \vspace{.5cm}

\end{titlepage}

\renewcommand*{\thefootnote}{\arabic{footnote}}
\setcounter{footnote}{0}

\thispagestyle{plain}

\Large
\begin{center}
\textbf{\papertitle}

\vspace{1cm}

\end{center}
\normalsize
\noindent
\textbf{Abstract:} We study how sophistication in generative AI (genAI) use varies among the back-office workforce of a large firm. Using proprietary data, we observe 713,564 employee prompts and their corresponding large language model responses from nearly 4,000 back-office employees across 15 functional areas over eight months in 2025. We document three main findings. First, senior employees exhibit more sophisticated genAI use, consistent with domain expertise complementing genAI capabilities. Second, sophistication varies considerably across functions and is highest in Strategy, Digital Innovation, and Project Management, three groups that share a focus on firmwide strategic initiatives and organizational change. Third, we observe neither improvements in sophistication over time nor lasting improvements following formal AI training, suggesting that sophisticated use can be difficult to change. Together, our study provides measures of and insights into sophisticated genAI use that managers can use to improve outcomes and that researchers can use in future research.

\vspace{1cm}
\begin{flushleft}


\vspace{5mm}

\end{flushleft}

\newpage
\pagenumbering{arabic}
\normalsize
\singlespacing

\section{Introduction} \label{section:Intro}

Companies have rapidly adopted generative artificial intelligence (genAI), spending an estimated \$37 billion in 2025 \parencite{T.R.D+2025}. 
Yet adoption alone does not guarantee a return on investment.
As companies face costs and token-based usage limits, ``encouraging the right behaviors to get the most out of every AI interaction'' is going to be necessary to achieve organizational goals \parencite{C.D.D+2026adaptation}.
To provide insights into how people interact with genAI, we examine how genAI sophistication varies over time, across seniority levels and functional areas, and following AI training within a large firm that has already achieved widespread genAI adoption.

We define sophisticated genAI use as skilled use informed by an understanding of how the tool works and where it can be applied. 
Such use is reflected in clear and specific instructions, deliberate prompting techniques, and application of the tool across a broad mix of tasks. 
While a growing literature has examined whether, how often, and for what purposes genAI is used \parencite{BickBlandinDeming2026, C.C.D+2025, H.T.M+2025, C.C.D+2026, B.B.D+2026microstructure}, these studies provide little evidence on how employees interact with genAI and, more importantly, how well they do so.
We extend this literature while also providing insights that managers can use to better measure AI use within their organizations.

We study variation in genAI sophistication within the back-office group of KPMG LLP (KPMG), which provided its employees with access to multiple genAI tools during the eight-month period from January through August 2025: including Microsoft Copilot and aIQ Chat, an enterprise-specific chat tool that provided access to commercial large language models (LLMs) such as those from Anthropic and OpenAI. 
For aIQ Chat, we observe the full conversation transcripts. 
To measure genAI sophistication, we prompt an LLM to classify each conversation by use case and assess the clarity and specificity of its instructions and the prompting techniques employed, producing conversation-level measures of the dimensions of sophisticated use.
For Microsoft Copilot, we observe how frequently employees use the tool but not the content of their conversations. 
We aggregate our conversation-level measures and Copilot usage to the employee-month level and link them to each employee's seniority, functional area, and completed genAI training. 
Our sample includes employees in IT, marketing, accounting, finance, and other non-client-facing administrative functions. 
Because comparable functions exist in most large organizations, our setting should provide evidence relevant to managers in companies beyond KPMG.

We begin by documenting employee adoption and usage frequency. On average, 83.9\% of employee-months include use of at least one of the two genAI tools, with Copilot used more frequently than aIQ Chat.
This rate exceeds the 32.1\% reported in prior survey evidence \parencite{BickBlandinDeming2026}, possibly because the firm we study strongly encourages its employees to use genAI.
Broad adoption, however, masks substantial variation in usage intensity. 
Among employees who use aIQ Chat during the sample period, the top decile accounts for more than half of all conversations. 
Senior employees use the tools more frequently, primarily because of greater Copilot use. 
Monthly adoption rates also vary across functional areas, from 73\% in Compliance to 94\% in Internal Audit.
Finally, usage changes modestly over the eight-month window, consistent with the firm having reached a relatively mature stage of adoption.

We next document the tasks for which employees use aIQ Chat. 
These use-case categories form the basis for our subsequent measure of how broadly each employee applies the tool. 
Writing-related tasks, including requests to edit user-provided text, generate new content, or both, appear in roughly three-quarters of conversations. 
Employees also use the tool for knowledge retrieval, document understanding, and software guidance, with smaller but meaningful proportions devoted to coding and data analysis, ideation, and other tasks.
Use cases also vary across seniority levels and functional areas. 
More senior employees make more knowledge-retrieval requests, whereas staff employees make more writing and personal requests. 
Communications and Sales \& Marketing employees seek more assistance with writing and ideation, while Information Technology employees make more requests involving coding and software guidance.
Together with the adoption analyses, these findings provide context for our analysis of how sophistication varies within the firm.

Against this backdrop, we turn to the paper's primary analyses and examine variation in sophisticated use. 
To operationalize this concept, we use three composite measures: use-case diversity, deliberate strategy use, and prompt clarity. 
Each measure corresponds to one component of sophisticated use: applying genAI across a broad mix of tasks, employing deliberate prompting techniques, and providing clear and specific instructions. 
Our analyses of variation across employee-months yield three main findings. 
First, we do not observe employees' genAI use becoming more sophisticated over the eight-month sample period. 
Second, each measure rises with seniority, with staff ranking below managers and managers below employees above the manager level. 
Third, sophistication varies considerably across functional areas. 
It is highest in Strategy, Digital Innovation, and Project Management, three groups that share a focus on firmwide strategic initiatives and organizational change.
Accounting \& Finance lags on all three sophisticated use measures.

Our sophistication measures require conversation transcripts, which firms rarely retain and are costly to process.
We therefore examine whether these measures relate to four usage characteristics that managers can observe from metadata alone: how much an employee asks the tool to do (ambition), how much the employee iterates on a request (persistence), how often the employee uses aIQ Chat (frequency), and how often the employee also uses Copilot, another of the firm's genAI tools (flexibility).
A reliable association would give managers a low-cost way to gauge how well employees use genAI, not only how often.
In simple correlations, regressions with additional controls, and models with employee fixed effects, we consistently find that greater ambition and persistence, reflected in longer initial prompts and more iteration within conversations, accompany more sophisticated use. 
Finally, we examine whether formal AI training is associated with increases in employee genAI sophistication. 
In models with employee fixed effects, we find more sophisticated use in the month an employee completes formal genAI training but not in the months that follow, suggesting that any improvement associated with these trainings is temporary.
While our evidence does not establish that the training was ineffective, it provides little evidence that completion of formal training produces lasting changes in observed behavior.

Our study provides managers with several practical benefits.
First, we provide a customizable process for measuring sophisticated genAI use within their organizations.
Second, we provide descriptive evidence from a firm with widespread adoption.
We do this across employee groups, in relation to easily observable measures, and across periods before and after formal AI training.
This evidence creates genAI benchmarking opportunities that can help managers evaluate their organization's progress and inform resource allocation decisions.
For example, the absence of improvement over time or sustained improvement following training illustrates the difficulty organizations may face in fostering lasting changes in employee genAI use.

Our paper also makes several contributions to the academic literature. 
Existing research documents rapid genAI adoption, common workplace use cases, and productivity effects of access to particular AI tools \parencite[e.g.,][]{BickBlandinDeming2026, A.B.B+2025, W.G.A2024, B.B.D+2026microstructure, C.C.D+2026, C.X2026}.
Yet this research does not evaluate how well employees use genAI. 
We add to the literature by introducing new measures of sophisticated use and by providing field evidence on how sophisticated genAI use varies among employees inside one organization after adoption is already widespread. 
We find that sophistication rises with seniority and differs substantially across functional areas. 
We also observe little systematic improvement over time. 
Sophistication is higher in the month of formal training but not in subsequent months. 
Together, these findings help us better understand the evolving genAI adoption timeline in which attention shifts from whether employees adopt genAI to how sophisticated use is distributed and changes after adoption.

\section{Background and Literature Review} \label{section:Related}

\subsection{From Adoption to Sophistication}
Companies have rapidly adopted genAI.
In a single year, employee access to AI tools has expanded by 50 percent \parencite{C.D.D+2026adaptation}.
Despite its rapid adoption, recent and contemporaneous research documents uneven workplace diffusion across workers, firms, business functions, and tasks \parencite{BickBlandinDeming2026, A.B.B+2025, B.B.D+2026microstructure, H2026, H.V2025}.
For example, a working paper by \textcite{A.B.B+2025} document that while 64 percent of the employees they surveyed report using AI, only 20 percent report using it frequently.
\textcite{BickBlandinDeming2026} show that the intensity of AI use and related productivity gains vary by industry and firm climate.
Similarly, a working paper by \textcite{B.B.D+2026microstructure} shows that within adopting firms, use can vary substantially by function.

Frequent use is a necessary precursor to achieving organizational benefits.
For example, using a large randomized experiment of 6,000 workers at 56 firms, \textcite{D.J.P+2025} show that genAI reduces time spent on email and accelerates document completion but only for the 40 percent of workers who used it frequently.
Similarly, \textcite{E.G.M2024} show in a lab experiment that genAI effects depend on whether managers and auditors both use and respond to AI-generated information.
Combined, these studies demonstrate that company genAI adoption by itself will not necessarily result in work benefits if employees' adoption consists of only infrequent use.

Large-scale analyses of AI conversations show that writing, information retrieval, analysis, and other forms of information work (e.g., the creation, processing, and communication of information) make up most of the common workplace uses of genAI \parencite{C.C.D+2025, C.C.D+2026, H.T.M+2025, T.J.W+2025}.
For example, a working paper by \textcite{C.C.D+2025} classifies a large sample of ChatGPT conversations and find that writing, practical guidance, and information seeking dominate AI use cases, especially in professional work.
Similarly, \textcite{H.T.M+2025} examine millions of Claude conversations and find that software development and writing dominate the use cases on that platform.
Each of these studies examines users on a single platform to reveal what people do with the tool, but generally not how skillfully individual employees do it.

Information systems research distinguishes use from effective use, emphasizing that organizational benefits depend on how users engage with a technology rather than merely whether they access it \parencite{O2000, B.G2013}.
How users engage matters for genAI, as variation in prompts meaningfully affects outcomes on specific tasks \parencite{C.G.G2026, E.S.V+2024, S.W2026}.
Across tasks, a recent report from AI lab Anthropic \parencite{S.B.L+2026} measures the ``AI fluency'' of a sample of its subscribers for a seven-day period, which the authors define as iterative, evaluative, and augmentative behaviors in AI conversations.
While they study associations among behaviors at a conversation level, we ask how sophisticated genAI use varies within an organization in which adoption is already widespread.
Answering this question requires measures of sophisticated use, which we develop from conversation transcripts in Section~\ref{section:Measures}.

\subsection{Seniority, Training, and Functional Area Differences}
We organize the remaining literature around three dimensions of potential variation in sophisticated use: seniority, formal training, and functional area.
Seniority partly reflects accumulated experience, and prior research provides mixed evidence on the relationship between experience and AI.\footnote{We acknowledge that we cannot reconcile the mixed findings from prior research.}
AI assistance can generate larger productivity gains for less-experienced workers by transmitting best practices \parencite{B.L.R2025, G.Q.S+2024}, yet domain expertise can improve the development, selective use, and evaluation of AI outputs \parencite{L.W2021, C.X2026}.
\textcite{W.G.A2024} show that the benefits of AI can be lower for senior workers when they exhibit lower trust in AI.
However, prior studies primarily estimate the productivity benefits of access to a particular AI application or tool, or they measure firm-level AI capability.
Instead, we examine whether seniority is associated with clearer, broader, and more deliberate genAI use.

Prior research identifies training as an important factor for achieving workplace AI adoption and perceived value.
For example, \textcite{H2026} shows that workplace training strengthens AI adoption.
\textcite{H.V2025} document that employees \emph{perceive} a need for training to achieve productivity gains from ChatGPT.
Practitioner evidence likewise emphasizes training as an important lever for converting genAI access into returns \parencite{K.P.T2025, Brown2025}.
These studies suggest that formal AI training should improve employees' ability to use genAI.
However, prior work does not establish whether such behavioral improvements from training actually occur.

AI adoption and use vary across business functions, occupations, and tasks, with especially prominent application in sales and marketing, strategy, information technology, writing, information retrieval, and document analysis \parencite{B.B.D+2026microstructure, C.C.D+2026, H.T.M+2025, T.J.W+2025}.
These patterns suggest that workflow fit and task composition influence not only whether employees use genAI, but likely how effectively they engage with it.
Therefore, we are likely to observe differences in sophisticated use across functional areas.

\section{Research Design} \label{section:Data}
\subsection{Setting and Data Sources}
We obtain proprietary data on the back-office workforce of a large professional services firm for January through August 2025.
Although these employees are not client-facing, they perform administrative functions critical to the organization's success, such as IT, internal audit, strategy, public relations, marketing, accounting, and finance.\footnote{Data for the Human Resources group were withheld because its conversations might discuss sensitive matters.}
Thus, our findings should be relevant beyond professional services firms to employees in comparable functions within other large organizations.

We observe two LLM channels that the firm provides to employees.
The first is a standalone chat interface called aIQ Chat that, during our sample window, provided users with access to commercial LLMs from several vendors, including Anthropic's Claude 3.5 Sonnet, OpenAI's GPT-4o and o1, as well as models from Google's Gemini, Meta's Llama, and others. 
GPT-4o was the default model. 
The transcripts for employees' interactions with aIQ Chat provide the basis for our primary analyses of sophisticated use.
The second LLM channel is Microsoft Copilot, an embedded assistant available through workplace software such as Outlook and Teams.

\subsection{Sample Construction}
We analyze the data at two levels:
\begin{enumerate}
\item \textbf{Conversation level:} Full transcripts of aIQ Chat conversations, including all user prompts and LLM responses within each conversation.
\item \textbf{Employee-month level:} Monthly use volume measures for both aIQ Chat and Copilot, plus aggregate measures derived from aIQ Chat transcripts.
\end{enumerate}
We link the conversation transcripts and frequency measures to employee seniority, functional area, and completion of firm-provided AI training.
As we do not observe the content of Copilot interactions, we use only aIQ Chat to construct conversation-level measures and examine sophisticated use.

Table~\ref{tab:sample_construction} summarizes the sample.
Panel A starts from 5{,}191 back-office employees.
Of these employees, 3{,}925 (75.6\%) use aIQ Chat at least once during any month of the sample window.
The Copilot dataset covers 4{,}392 employees, of whom 3{,}962 (90.2\%) use Copilot and 2{,}861 (65.1\%) use both tools.
Panel B reports the scale of aIQ Chat activity.
We observe 158{,}496 conversations and 713{,}564 user prompts, or 4.5 user prompts per conversation.
Usage is highly skewed.
The mean active employee has 40 conversations, while the median active employee has 14.
The top decile of active employees accounts for 51\% of conversations.\footnote{An ``active'' employee or employee-month has at least one aIQ Chat conversation.}

\section{Measures} \label{section:Measures}
This section describes how we examine aIQ Chat transcripts to create structured data and measures at the conversation and employee-month levels.

\subsection{Structuring Data}
We submit each aIQ Chat transcript to an LLM along with a metaprompt, as reproduced in Online Appendix~\ref{appendix:MetaPrompts}.
The metaprompt instructs the LLM to use structured coding rules to identify the conversation's use cases, prompting strategies, and other characteristics of sophisticated use.
We treat the resulting LLM outputs as structured measurement data.\footnote{We developed and validated the metaprompt through an iterative calibration procedure. In pilot rounds, the research team manually reviewed the LLM's outputs. We then ran duplicate API calls on the same transcripts and monitored one-sided positive classifications---cases in which only one run returned a given label, score, or flag---as a diagnostic of instability in the structured output. Disagreement cases were retained for further review and used to refine the coding instructions: we prompted the LLM to explain the basis for competing judgments and to identify ambiguities in the prompt language, while the research team retained final authority over any revisions. Through this process, we modified model choice and prompt wording, removed classifications that proved insufficiently reliable, added or revised classifications where needed, and compared candidate models before ultimately selecting OpenAI o1. All analyses were conducted under the firm's confidentiality requirements using de-identified data.}

\subsection{Conversation-Level Measures}
First, we classify each conversation by use case.
The use case categories consist of Writing \& Communication (\WRITINGCOMMUNICATION), Coding \& Data Analysis (\CODINGDATAANALYSIS), Text-Based Analysis (\TEXTBASEDANALYSIS), Knowledge \& Expertise (\KNOWLEDGEEXPERTISE), Software \& Tool Guidance (\SOFTWARETOOLGUIDANCE), Creative Thinking \& Ideation (\CREATIVETHINKINGIDEATION), Personal \& Non-work (\PERSONALNONWORK), and Other (\OTHER).
We also decompose each of these broad categories into subcategories.
We note that workplace conversations often blend tasks, such as summarizing a policy and drafting a stakeholder email in the same exchange. 
In such instances, the conversation is classified into multiple categories.

The same metaprompt scores several dimensions of either the first substantive user prompt or the overall conversation.\footnote{We use the first ``substantive'' prompt because employees occasionally begin conversations with a common greeting, such as ``hello.''}
For the first substantive prompt, ratings capture linguistic complexity (\INITIALLANGUAGECOMPLEXITY) and task and structural complexity (\INITIALCOMPLEXITY).
For the overall conversation, measures consider prompt organization and structure (\PROMPTSTRUCTUREQUALITY), overall prompting sophistication (\USERPROMPTINGSOPHISTICATION), specificity and precision (\SPECIFICITYSCORE), and clarity of the requested output format (\OUTPUTFORMATCLARITY).

We also obtain binary indicators for whether the user uploads a document (\DOCUMENTUPLOADED), specifies response constraints (\CONSTRAINTSPRESENT), requests structured output such as a table or JSON (\STRUCTUREDOUTPUTREQUESTED), provides acceptance criteria for the response (\ACCEPTANCECRITERIAPROVIDED), expresses explicit satisfaction with a previous response (\USERSATISFIED), makes frequent typos (\FREQUENTTYPOS), uses all-lowercase text (\ALLLOWERCASE), or includes pleasantries (\PLEASANTRIESUSED).
Finally, we code prompting strategies: role or persona assignment (\ROLEPLAY), few-shot examples (\FEWSHOTEXAMPLES), explicit requests for step-by-step reasoning (\CHAINOFTHOUGHT), self-verification or checking (\SELFVERIFICATION), and instructions for the model to ask clarifying questions (\INTERACTIVEREFINEMENT) \parencite{S.I.B+2024, W.W.S+2023, W.W.S+2023a, S.W2026}.

\subsection{Employee-Month Aggregation and Composite Measures} \label{section:CompMeasures}
We aggregate conversation-level measures to the employee-month level and restrict the sample to active employee-months, i.e., months when an employee has at least one aIQ Chat conversation.
As before, we define sophisticated genAI use as skilled use informed by an understanding of how the tool works and where it can be applied. 
Such use is reflected in clear and specific instructions, deliberate prompting techniques, and application of the tool across a broad mix of tasks.
To operationalize this definition, we use three composite measures that proxy for distinct but overlapping aspects of sophisticated use.
The first composite measure, prompt clarity (\CLARITYZ), measures how clearly and specifically users formulate requests.
We construct it by first combining standardized values of nine variables, including assessed organization, specificity, the presence of set constraints and requests for structured output, and then by standardizing the resulting composite measure.
Deliberate strategy use (\ANYSTRATEGY), our second composite measure, identifies how often the users actively employ prompting techniques in their conversations, such as role assignment or instructions for the LLM to self-verify.
Use-case diversity (\USECASEDIVERSITY), our third composite measure, identifies whether an employee applies aIQ Chat across a narrow or broad mix of task categories during the month.\footnote{See Appendix~\ref{appendix:VariableDefinitions} for a complete, detailed description of each measure.}
We interpret higher values of our composite measures as proxies for sophisticated use, not as direct measures of output quality or productivity.

\subsection{Presentation of Results} \label{section:presentation}
We present our results as dot-plot figures and report the underlying tables and statistics in Online Appendix~\ref{appendix:Tables}.
Each dot-plot reports one outcome measure (e.g., use cases), with rows that correspond to one particular group, i.e., a month, a seniority level, or a functional area.
A dot marks the group's mean value, plotted on a horizontal axis whose center reference line denotes the overall sample mean (or zero, in Figure~\ref{fig:bivar}); the axis is labeled with the sample mean at the center and the group minimum and maximum at the ends, so a dot's horizontal position shows how far the group lies above or below the sample average.
Color and fill jointly depict the direction and statistical significance of each group's difference-from-rest test using a scheme that is common for all of the dot-plot figures: a \emph{filled burnt-orange} dot depicts a mean value significantly above the sample mean, an \emph{open burnt-orange} dot depicts a mean value above the sample mean that is not statistically significant, a \emph{filled gray} dot depicts a mean value significantly below the sample mean, and an \emph{open gray} dot depicts a mean value below the sample mean that is not significant.
The tables in Online Appendix~\ref{appendix:Tables} report the exact values, sample sizes, and significance levels for all dot-plot figures.

\section{Extent of GenAI Use} \label{section:AdoptionResults}
As context for our primary analyses, this section first describes the prevalence and volume of genAI use in our sample and how use varies over time, seniority, and functional area, and then compares these patterns with evidence documented in prior research.

\subsection{Variation in Extent of GenAI Use} \label{subsection:AdoptionVariation}
Figure~\ref{fig:adoption} summarizes use volume for the employee-month panel as described in Section~\ref{section:presentation}. 
Usage is broad in our sample.
On average, 83.9\% of employee-months include use of at least one tool (\textit{Any Tool}).
At the employee-month level, Copilot use (78.0\%) is substantially more common than aIQ Chat use (43.7\%).

Panel A reports usage by month.
We do not observe rapid or accelerating growth during the January--August 2025 window.
Instead, the month-to-month changes are modest.
The proportion of users with any Copilot use rises over the sample window from a minimum of 71.0 percent in January to a maximum of 82.8 percent in August, while aIQ Chat usage is comparatively flat with most months not statistically different from the sample mean. 
This pattern is inconsistent with a simple story of uniformly increasing chat use over time.

Panel B shows that usage positively correlates with seniority.
Above-manager employees have higher overall usage driven by more Copilot use, while aIQ Chat usage is flatter across seniority groups.
Panel C reports variation across functional areas.
Digital Innovation, Communications, Sales \& Marketing, and Internal Audit are typically above the sample mean.
Accounting \& Finance is the only functional area whose average is significantly below the sample mean for every reported usage measure.
The contrast between Internal Audit and Accounting \& Finance is notable because these areas may draw on overlapping professional backgrounds.
In general, aIQ Chat and Copilot use are positively correlated across functional areas, but the relationship is not uniform.
Some functions appear to rely more heavily on one channel: for example, Specialized Services is above the sample mean for Copilot use but below the sample mean for aIQ Chat use, while Project Management shows the opposite pattern.

\subsection{Comparison with Prior Adoption Evidence} \label{subsection:AdoptionComparison} 
Our adoption rates are higher than rates reported in prior research \parencite{BickBlandinDeming2026, A.B.B+2025, H.V2025, H2026, B.B.D+2026microstructure}.
For example, \textcite{BickBlandinDeming2026} survey U.S. workers in August and November 2024, shortly before our January 2025 sample begins, and report that 32.1\% of employed respondents used genAI for work and 27.3\% used it at work in the prior week.
In our setting, 83.9\% of employee-months use at least one enterprise LLM channel.
There are several possibilities for this difference.
First, because we study an organization that provides and encourages use of genAI tools, our study, by construction, omits individuals who lack this organizational support.\footnote{Table 1, column 3, in \textcite{BickBlandinDeming2026} reports a large and positive relationship between employee AI use and employer encouragement.}
Second, \textcite{BickBlandinDeming2026} report especially high work adoption in computer and mathematical occupations (53.6\%), management occupations (52.0\%), and business and finance occupations (48.2\%).
Our sample of office workers disproportionately captures these higher use groups while omitting lower use groups like personal services (15.8\%).
Although \textcite{BickBlandinDeming2026} conducted their survey at an earlier point in the adoption curve than our January--August 2025 analyses, our lack of a strong upward time trend suggests that timing plays a limited role in explaining the difference.
Finally, while self-reported adoption rates in surveys may understate actual use \parencite{L.K.I2026}, we observe proprietary usage logs and complete employee genAI conversation transcripts. 

Our seniority finding contrasts both with conventional wisdom and, to a lesser extent, with prior academic research. 
Conventional wisdom holds that recent entrants, already fluent in these tools, will out-perform experienced colleagues; executives say so directly and young degree-holders report the highest confidence in their own AI readiness \parencite{PohleFernandez2026AINatives}. 
Survey evidence already contradicts that view: \textcite{BickBlandinDeming2026} find work use follows an inverted U in age, peaking in users' 30's and 40's rather than among the youngest workers. 
Our results reinforce that correction at the young end but diverge at the other, with use rising monotonically rather than turning down among senior employees. 

Our functional area analyses somewhat echo occupation-based evidence.
For example, while \textcite{BickBlandinDeming2026} report high work adoption in computer and mathematical, management, and business and finance occupations, we report high usage in the computer-focused area Digital Innovation and among employees who manage others (as part of our seniority tests).\footnote{We do not identify math-focused personnel.}
We further disaggregate business-focused areas and report heterogeneity that resembles evidence from \textcite{B.B.D+2026microstructure}.
They report firm-level evidence from the November 2025--January 2026 Business Trends and Outlook Survey and find that, among firms reporting AI use in at least one business function, Sales \& Marketing is the most common function for use, followed by Strategy \& Business Development, then Information Technology.
This ordering resembles our finding of relatively high usage in Sales \& Marketing, Strategy, and Digital Innovation. 
While their employment-weighted results place Finance \& Accounting among the leaders, we find the opposite. 
The difference need not represent a contradiction as their measure identifies whether firms deploy any form of AI within a function, while ours measures how frequently employees in that function use two enterprise genAI tools.
Taken together, inferences drawn from our sample are broadly consistent with large-scale survey evidence.

\section{GenAI Use Cases} \label{section:UseCaseResults}
As context for our primary analyses, this section first describes the genAI use cases in our sample and how use varies over time, seniority, and functional area, and then compares these patterns with prior use-case evidence.

\subsection{Variation in GenAI Use Cases}
Table~\ref{tab:use_cases} reports our categorized use cases at the conversation level.\footnote{Recall that we allow multiple use-case labels for a single conversation.}
Writing appears in 73.3\% of conversations, far exceeding any other category, and includes subcategories for editing user-provided material (45.8\%) and generating new text (34.8\%). 
Employees also use aIQ Chat to retrieve, interpret, and apply information.
Knowledge/expertise queries appear in 23.0\% of conversations, and text/document analysis appears in 10.7\%.
Software/tool guidance appears in 9.9\% of conversations.
Coding/data analysis appears in 7.3\% of conversations.
Ideation and personal requests appear less often but still represent nontrivial proportions.

Figure~\ref{fig:use_cases_by_group} reports the proportion of conversations assigned to each use case by month, seniority, and functional area, as described in Section~\ref{section:presentation}.\footnote{The unit of analysis for the underlying use-case labels is the \emph{conversation}; group-level statistics summarize the proportion of conversations in the group assigned each label.}
Across months in Panel A, we do not observe a clear monotonic shift in how employees use aIQ Chat.
Writing remains the baseline use case throughout the sample window.
Secondary categories move modestly: coding/data analysis increases somewhat, while knowledge retrieval and ideation decrease.

Panel B shows that use cases vary by seniority.
Staff conversations skew toward writing and personal requests.
Managers make relatively more coding and data-analysis requests.
Above-manager employees are the most knowledge-intensive users, which is the largest swing across seniority groups in Panel B.
This pattern is consistent with employees using LLMs in ways that reflect their responsibilities and complement their domain knowledge.

Panel C suggests that use case patterns align with expected workflow differences across functional areas.
Administrative Services and Internal Operations show the highest proportion of personal requests, consistent with work tasks that may appear personal, such as travel and scheduling performed by these groups.
Information Technology has comparatively more coding and tool guidance.
Communications and Sales \& Marketing show elevated ideation alongside heavy writing.
Taken together, these patterns show that writing is common across the organization, while secondary use cases differ meaningfully by functional area.
We observe that functional areas with a higher concentration of text-based analysis and knowledge use also tend to have higher overall usage.

\subsection{Comparison with Prior Use Case Evidence}
Our use-case evidence aligns with an emerging consensus that writing is the leading workplace use of genAI \parencite{BickBlandinDeming2026, C.C.D+2025, H.T.M+2025, C.C.D+2026}.
Although taxonomies differ, our other top use cases resemble common categories reported in other studies.
For example, our second most prominent use case, Knowledge \& Expertise, shares characteristics with searching for facts or information in \textcite{BickBlandinDeming2026}, seeking information in \textcite{C.C.D+2025}, or information retrieval in \textcite{C.C.D+2026}.
As for variation in use, \textcite{C.C.D+2025} find that work-related ChatGPT messages from management and business users are especially writing-heavy, while messages from computer-related users request more technical help. 
We find similar results as writing is prevalent at all levels and among all business users, although less so in Information Technology, where we find higher rates of technical assistance (which we label \SOFTWARETOOLGUIDANCE).
Despite examining a single organization, these comparisons suggest that inferences drawn from our sample are broadly consistent with evidence from surveys and analyses of specific platforms.

\section{Sophistication of GenAI Use} \label{section:PromptResults}
We next report the paper's primary analyses regarding sophisticated use.
We first report summary statistics and then examine how sophisticated use varies over time, seniority, and functional area.
Next, we examine how sophisticated use relates to training and easily observable behaviors, and finally we compare our findings to prior related evidence.

\subsection{Sophisticated Use Summary Statistics} \label{subsection:PromptSummaryStats}
Table~\ref{tab:style_technique} summarizes LLM-assigned prompt input-output measures, binary prompt features, prompting strategies, and the sophisticated use measures \CLARITYZ, \ANYSTRATEGY, and \USECASEDIVERSITY.\footnote{Section~\ref{section:Measures} and Appendix~\ref{appendix:VariableDefinitions} describe how we construct these measures.}
The first six prompt input-output variables are scored 1 to 5, and all show meaningful variation in these assessed usage measures.
The next three variables are binary measures, as follows.
Explicit constraints (\CONSTRAINTSPRESENT) appear in 24.3\% of conversations, but structured-output requests (\STRUCTUREDOUTPUTREQUESTED) appear in only 2.9\% of conversations, and explicit acceptance criteria (\ACCEPTANCECRITERIAPROVIDED) appear in only about 1.0\% of conversations. The variable \CLARITYZ{} is a composite constructed from these nine variables listed above it.

Deliberate prompting strategies are uncommon, as follows.
Role prompting (\ROLEPLAY) occurs in 4.4\% of conversations.
Few-shot examples (\FEWSHOTEXAMPLES), explicit chain-of-thought requests (\CHAINOFTHOUGHT), self-checking prompts (\SELFVERIFICATION), and interactive refinement (\INTERACTIVEREFINEMENT) each appear in less than 1\% of conversations.
The portion of conversations that use any of these strategies (\ANYSTRATEGY) is 5.2\%.

The inputs to our use-case diversity measure (\USECASEDIVERSITY) appear in Table~\ref{tab:use_cases}. 
Other measured prompt characteristics describe the user's interaction style, as follows.
Users have a conversational tone that includes pleasantries (\PLEASANTRIESUSED) in 30.2\% of conversations.
They communicate casually with frequent typos (\FREQUENTTYPOS) in 10.7\% of conversations and use entirely lowercase writing (\ALLLOWERCASE) in 5.4\% of conversations.
Users explicitly express satisfaction with a response (\USERSATISFIED) in 3.5\% of conversations and upload or reference documents (\DOCUMENTUPLOADED) in 5.5\% of conversations.
We report these interaction style measures solely as descriptive context.

\subsection{Variation in Sophisticated Use} \label{subsection:EvaluationCorrelates}
Figure~\ref{fig:evaluation} summarizes variation in our employee-month composite measures of sophistication: prompt clarity (\CLARITYZ), deliberate strategy use (\ANYSTRATEGY), and use-case diversity (\USECASEDIVERSITY).
See Section~\ref{section:presentation} for a description of the presentation style we use for this figure. 
Panel A suggests some potential improvement in our sophisticated use measures over time, but we interpret this as a consequence of a weak January rather than a sustained upward trend.
Panel B shows that more senior employees are more likely to use genAI for a broader set of use cases, with greater prompt clarity and a greater likelihood to utilize prompting strategies. 
Each composite measure rises with seniority: staff employees have the lowest values, managers have higher values, and above-manager employees have the highest values. 
Combined with the adoption patterns in Sections~\ref{subsection:AdoptionVariation} and~\ref{subsection:AdoptionComparison}, these results suggest that more senior employees use genAI more frequently and with greater sophistication. 

Panel C shows that sophisticated use also varies across functional areas with some interesting patterns.
Specifically, Project Management is the only area where all three measures are significantly above the sample mean.
Both Communications and Sales \& Marketing have above-mean prompt clarity and strategy use, but below-mean use case diversity. 
Figure~\ref{fig:use_cases_by_group} also reports higher rates of writing and ideation use cases for these groups, tasks that may align especially well with the available LLMs. 
This alignment may contribute to both the relatively sophisticated interactions observed here and the above-mean use volume reported in Figure~\ref{fig:adoption}.

We observe that both Digital Innovation and Strategy are above the mean for prompt clarity and use-case diversity.
These results may suggest the importance of attitudes: membership in the Digital Innovation group may reflect enthusiasm for new tools, while membership in the Strategy group may reflect comfort with uncertainty.
In a similar vein, Information Technology has above-mean strategy use and use-case diversity but below-mean prompt clarity, suggesting a greater technical focus on the tool.

By contrast, Administrative Services, Property \& Facilities, and Internal Operations are below the sample mean across the three sophisticated use measures.
Accounting \& Finance is also below the sample mean across all three measures, and is additionally distinctive because it is the only functional area significantly below the sample mean across all use-volume categories in Figure~\ref{fig:adoption}.
This pattern could reflect differences in user characteristics, department leadership, or the possibility that the available LLM tools were less naturally useful for spreadsheet-intensive accounting and finance tasks.

\subsection{Characteristics Associated with Sophisticated Use}
Companies rarely preserve transcripts indefinitely, and processing them is costly even when they are available.
Figure~\ref{fig:bivar} therefore relates the three sophisticated use measures to usage characteristics that firms can observe from metadata alone, and to completion of formal AI training.
We group these usage characteristics into four constructs.
Ambition is the length of the user's first prompt, on the logic that saying more indicates asking for more.
Persistence is the amount of iteration within a conversation, which reflects a user refining a request rather than accepting the first response.
Frequency is how often an employee uses aIQ Chat, which captures exposure to the tool rather than skill in using it.
Flexibility is how often an employee also uses Copilot, the firm's other genAI tool.

Without user fixed effects (Panel A), all three measures are higher when first prompts are longer (\UMAVGFIRSTUSERCHARS) and conversations include more iteration (\textit{ROUNDS}).
The three measures are also higher with contemporaneous training (\textit{TRAIN MO}), following cumulative past trainings (\textit{PAST TRAIN}), and when Copilot usage is higher (\LOGCOPILOTUSAGEDATES).
More frequent aIQ Chat use (\LOGUMNUMCONVERSATIONS and \LOGGPTUSAGEDATES) accompanies greater prompt clarity and use-case diversity, but less strategy use.

In Panel B, we absorb time-invariant user traits by including user fixed effects.
When we do this, we no longer see broad evidence that our measures of sophisticated use improve with Copilot use.
Observing an effect would have suggested that prompting across multiple genAI platforms generates learning that promotes greater sophistication.
Instead, the difference across Panels A and B suggests that those who demonstrate a capacity for more sophisticated use also choose to use multiple platforms.
Likewise, the null (and sometimes weakly negative) result for past training suggests that the positive association in Panel A reflects selection into training by more skilled users, rather than training-induced gains in sophisticated use. 
Overall, we do not find evidence that completing training leads to lasting, observable improvement in a user's sophistication.
We do, however, continue to see consistently positive associations with longer first prompts, more iteration, and training completed in the current month.

Table~\ref{tab:multivar} estimates multivariate associations by including multiple predictors simultaneously and comparing specifications with and without user fixed effects.\footnote{We omit \LOGUMNUMCONVERSATIONS{} from these specifications for its conceptual overlap with \LOGGPTUSAGEDATES.}
Across specifications, the same three variables remain consistently and positively associated with our sophisticated-use measures: longer first prompts, more iteration, and training completed in the current month (\UMAVGFIRSTUSERCHARS, \LOGUMAVGUSERROUNDS, and \LOGTRAININGSCOMPLETEDMONTH, respectively).
The contrast between current- and prior-month training (\LOGTRAININGSCOMPLETEDSTARTMONTH) is somewhat ambiguous: it may reflect a temporary boost while training is top-of-mind, or it may simply capture employees trying out the techniques during training.

Taken together, Figure~\ref{fig:bivar} and Table~\ref{tab:multivar} do not show reliably positive associations between our sophisticated use measures and either Copilot use or frequent use of aIQ Chat.
The reported associations do suggest that the sophisticated use measures are most tied to ambition and persistence.
Because first prompt length and iteration can be measured directly from conversation logs without prompting an LLM to classify conversation content, these measures may offer managers relatively low-cost signals of more sophisticated use.\footnote{We view these associations as useful low-cost signals for our sophisticated use measures, subject to the following considerations. The relationship between these measures and our sophisticated use measures is partly mechanical because longer and more iterative conversations create more opportunities to clarify intent or desired output, span multiple use cases, and employ a prompting strategy.}

\subsection{Comparison with Prior Evidence}
To our knowledge, only a small number of studies use conversation data to examine dimensions related to the quality of genAI use.
Two current working papers characterize interaction type or tool efficacy rather than variation in how skillfully users interact with genAI.
\textcite{C.C.D+2025} classify user messages to ChatGPT as ``asking,'' ``doing,'' or ``expressing.''\footnote{``Asking'' messages seek information or advice; ``doing'' messages request output created primarily by the model; and ``expressing'' messages neither seek information nor ask the model to perform a task.}
They find that asking messages are associated with greater apparent user satisfaction than doing or expressing messages under both an automated classifier and direct user feedback.
\textcite{T.J.W+2025} use LLM-assessed task completion and users' thumbs-up/down feedback to evaluate Copilot's efficacy across work activities.
They find that Copilot performs best for communicating, teaching or explaining, and writing, and worse for image generation and data analysis.

By contrast, a report by Anthropic examines observable behaviors intended to capture genAI fluency within individual conversations \parencite{S.B.L+2026}.
Specifically, they measure eleven conversation-level behaviors, including clarifying goals, specifying output formats, providing examples, checking facts, and iteratively refining outputs.
One of their principal findings parallels ours: conversations that include iteration and refinement exhibit more of the other deliberate and evaluative behaviors.
Our study differs in two important respects.
First, Anthropic analyzes conversations from a single seven-day window, whereas we follow employees over eight months.
This longer panel allows us to examine changes over time and, contrary to expectations, we find little evidence of sustained improvement.
Second, Anthropic's analysis does not link conversations to user or organizational characteristics.
We link conversations to employee seniority, functional area, and training records, allowing us to examine differences across employees and changes within employees following formal training.

\section{Discussion and Conclusion} \label{section:DiscussionConclusion}

In our study, we examine genAI use and sophistication among the back-office employees of a large firm that has already achieved widespread genAI adoption.
Employees have access to similar tools and organizational support, allowing us to study variation within a relatively common environment. 
We document how genAI use and sophistication vary over time and across seniority levels and functional areas, how they relate to easily observable behaviors, and how they change following firm-provided AI training. 
We provide four primary insights, which we discuss here.

First, senior employees exhibit more sophisticated genAI use.
This result is consistent with domain expertise complementing genAI capabilities, but several explanations are possible.
Senior employees may have greater delegation proficiency and therefore be more practiced at specifying objectives, reviewing intermediate work, providing feedback, and pushing for the output they want.
Alternatively, senior employees may also have stronger incentives to clarify requests and verify responses because they face greater accountability for the resulting work product.
These proposed explanations each suggest that sophisticated genAI use may reflect broader professional experience and not just genAI expertise or familiarity.

Second, sophistication varies substantially across functional areas.
Strategy, Digital Innovation, and Project Management exhibit the highest sophistication across our measures.
These three groups share a focus on firmwide strategic initiatives and organizational change, which may provide employees with both a broad set of potential applications and work that benefits from clearly defining and iterating on tasks.
By contrast, Accounting \& Finance exhibits below-average sophistication across all three measures and is also the only functional area significantly below the sample mean across all reported usage measures.
These group differences may reflect task mix, workflow fit, employee characteristics, department leadership, or attitudes toward genAI.
For example, \textcite{C.X2026} document productivity and reporting-quality gains for accountants using genAI built into an accounting workflow.
This suggests that a standalone chat interface may be less naturally suited to the spreadsheet-intensive and well-established processes common in Accounting \& Finance than to the evolving and open-ended tasks associated with organizational change.
Together, these results suggest that sophisticated use may depend not only on employee capabilities, but also on whether the work calls for the clear instructions, deliberate strategies, and broad application captured by our measures.

Next, we do not observe clear improvements in sophisticated use over the eight-month sample period.
While we find increased sophistication in the month when an employee completes formal AI training, we do not find increased sophistication in the months that follow.
While our evidence does not establish that the training was ineffective, it provides little evidence that completion of formal training produces lasting changes in observed behavior.
Together, the time and training results convey the difficulty organizations may face in producing sustained changes in sophisticated employee genAI use.

Finally, we find that longer initial prompts and more iteration within conversations accompany more sophisticated use.
In our analyses, these two behaviors are more consistently associated with sophisticated use than usage frequency.
While these relatively easy-to-track behaviors may provide useful signals when detailed transcript-based measures are unavailable, managers should not interpret any of these three behaviors as direct measures of performance.

Our analysis is descriptive and has important limitations.
While our composite measures proxy for sophisticated use, we are not able to determine whether the LLM output satisfied the user's goals.
Further, we are not able to directly measure output quality or genAI's influence on productivity.
Next, in the time since our window concluded, the field has continued its rapid advance.
Model developments have shifted best practices for prompting, and advancements in AI agents have likely changed the potential utility of genAI for use cases such as coding.
Consequently, the rates of use and prevalence of particular behaviors reported in this paper may already differ from current practice. 

Despite these limitations, companies are likely to benefit from employees who exhibit good AI behaviors and habits regardless of the existing AI technology. 
In addition, documenting evidence at a point in time is necessary to create reference points and historical benchmarks in a field that will continue to change. 
Further, while the paper's descriptive design does not offer causal inference, it can motivate research designs that will.
For these points, future research will be necessary to build knowledge of genAI's entry into, and its effects on, the workplace.

\clearpage
\singlespacing
\newrefcontext[sorting=nyt]
\section*{References}
    \printbibliography[heading=none]
    \singlespacing

\clearpage

\clearpage
\begin{landscape}
\begin{figure}[!htbp]
\centering
    \caption{Usage and Intensity}
    \label{fig:adoption}

    \textit{Panel A: By Month}\par\vspace{0.25em}
    \includegraphics[width=0.98\linewidth]{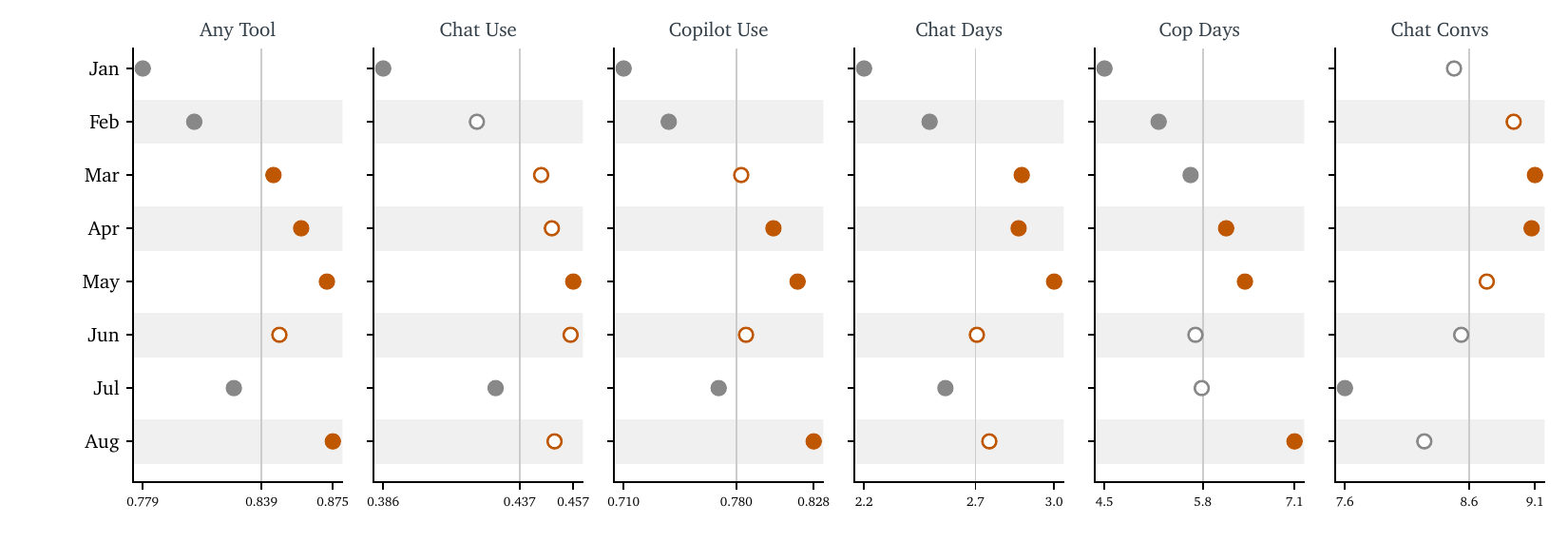}

    \vspace{1.0em}
    \textit{Panel B: By Seniority}\par\vspace{0.25em}
    \includegraphics[width=0.98\linewidth]{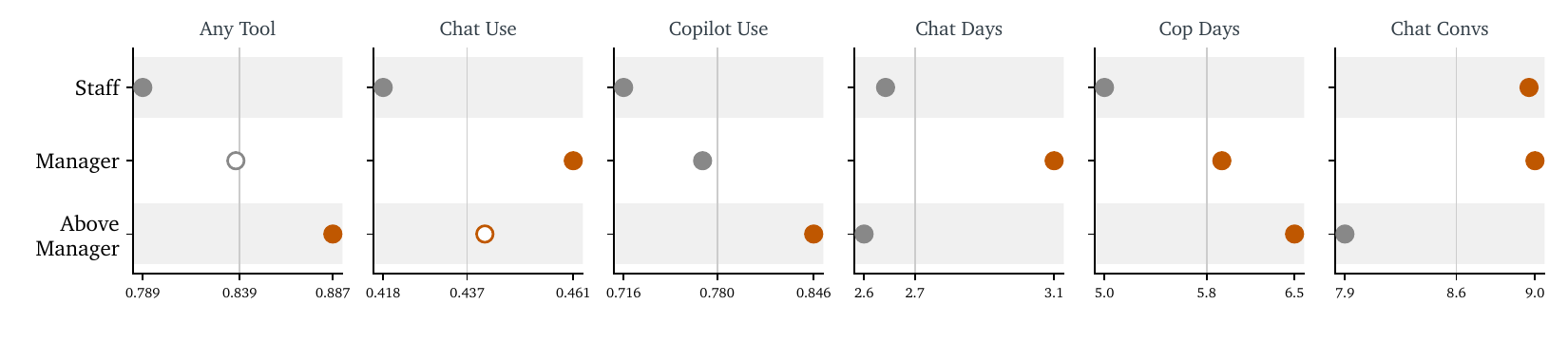}
\end{figure}
\end{landscape}

\clearpage
\begin{landscape}
\begin{figure}[!htbp]
\ContinuedFloat
\centering
    \caption[]{Usage and Intensity (continued)}

    \textit{Panel C: By Functional Area}\par\vspace{0.25em}
    \includegraphics[width=0.98\linewidth]{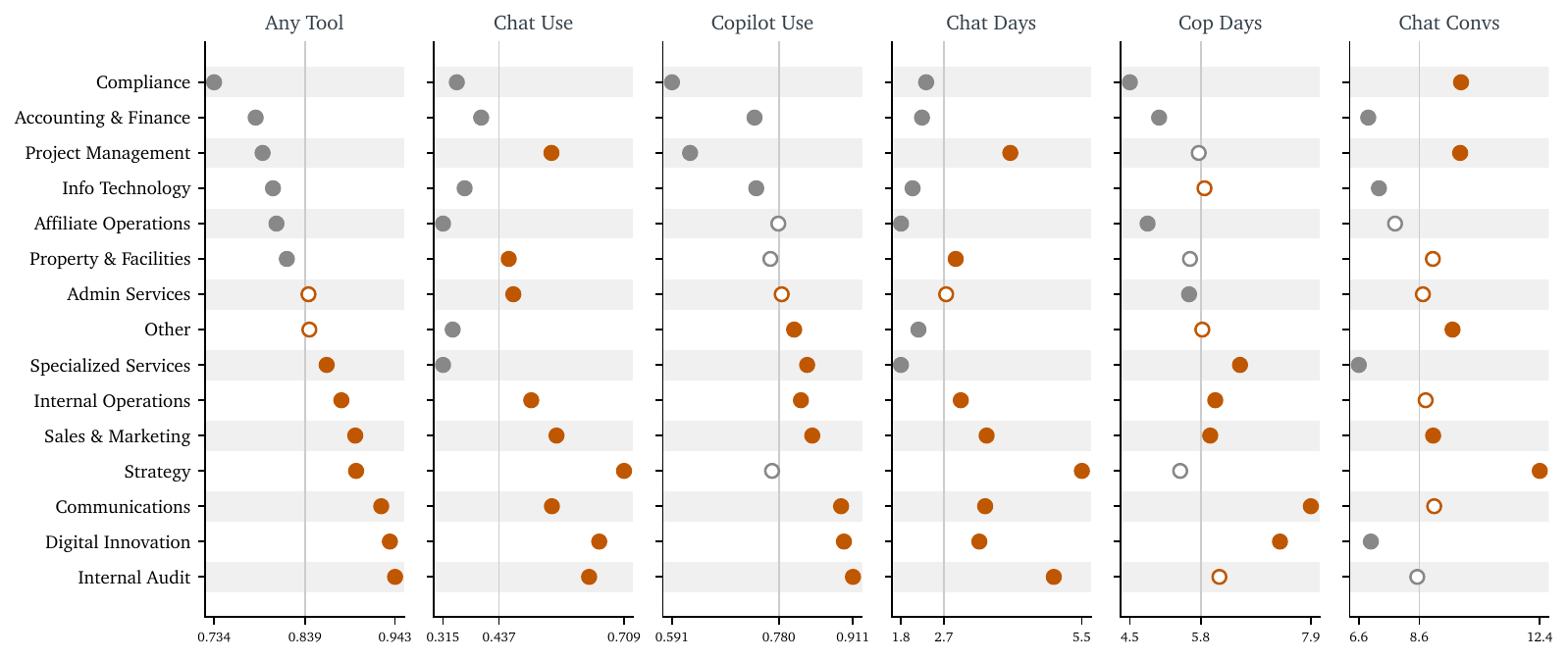}

    \floatfoot{\footnotesize\notesjustified \textit{Notes:} Each panel plots an employee-month usage measure (columns) for the groups listed---months in Panel~A, seniority levels in Panel~B, and functional areas in Panel~C. Each dot marks a group mean positioned relative to the overall sample mean (the vertical reference line); horizontal-axis ticks report the sample mean at the center and the group minimum and maximum at the ends, so a dot's horizontal position shows how far the group lies above or below the sample average. A dot's color and fill summarize the group-by-group difference-from-rest test: a filled burnt-orange dot denotes a group mean significantly above the sample mean, an open burnt-orange dot a value above the mean that is not statistically significant, a filled gray dot a value significantly below the mean, and an open gray dot a value below the mean that is not significant. Underlying values, sample sizes, and two-sided $p$-values are reported in Online Appendix Table~\ref{tab:adoption}. Variable definitions are provided in Appendix~\ref{appendix:VariableDefinitions}.}
\end{figure}
\end{landscape}

\clearpage
\begin{landscape}
\begin{figure}[!htbp]
\centering
    \caption{Use Cases by Group}
    \label{fig:use_cases_by_group}

    \textit{Panel A: By Month}\par\vspace{0.25em}
    \includegraphics[width=0.98\linewidth]{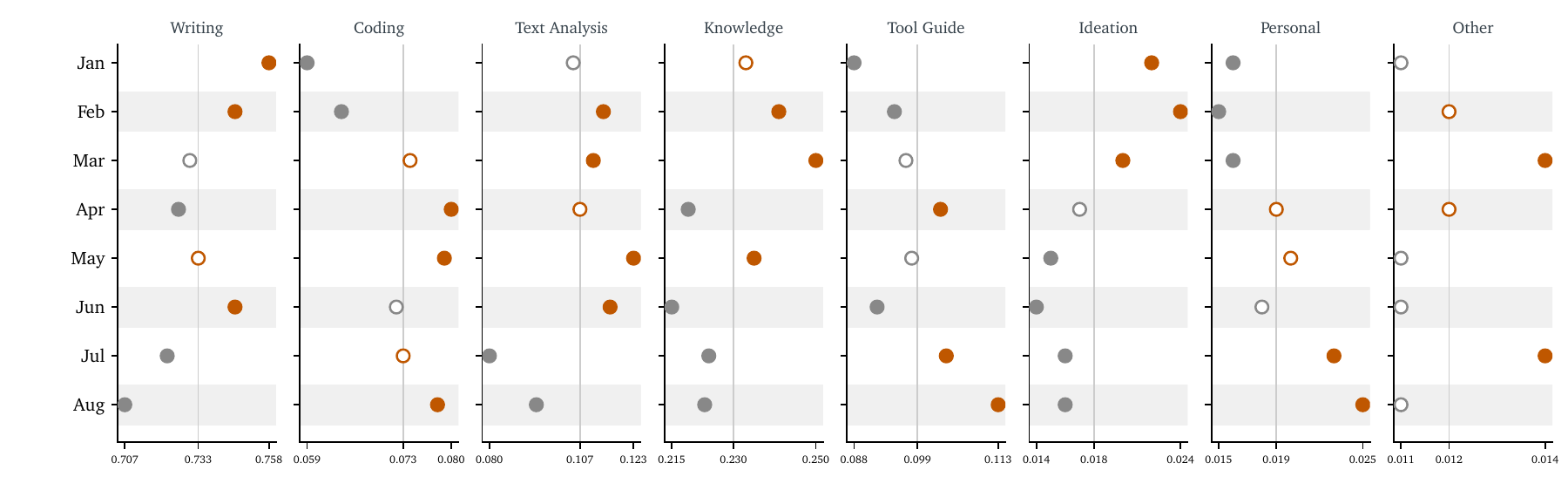}

    \vspace{1.0em}
    \textit{Panel B: By Seniority}\par\vspace{0.25em}
    \includegraphics[width=0.98\linewidth]{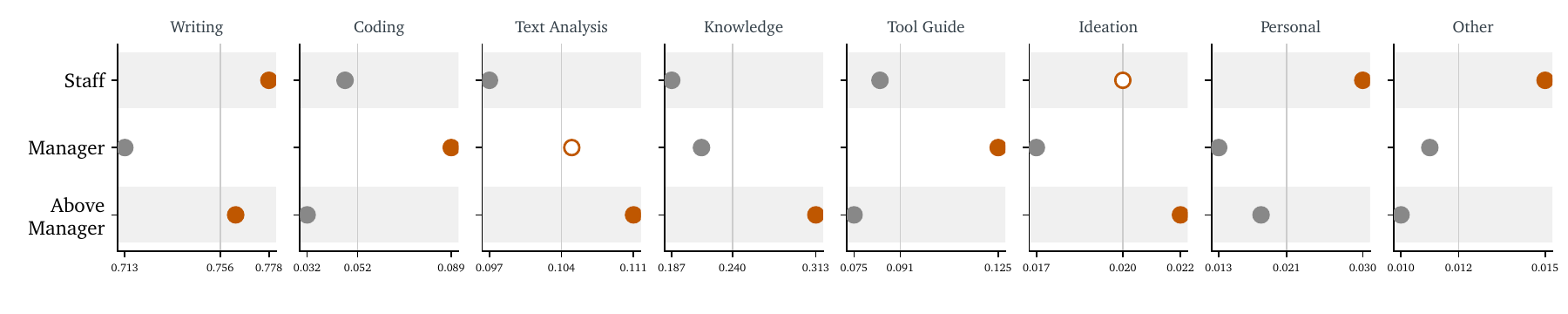}
\end{figure}
\end{landscape}

\clearpage
\begin{landscape}
\begin{figure}[!htbp]
\ContinuedFloat
\centering
    \caption[]{Use Cases by Group (continued)}

    \textit{Panel C: By Functional Area}\par\vspace{0.25em}
    \includegraphics[width=0.98\linewidth]{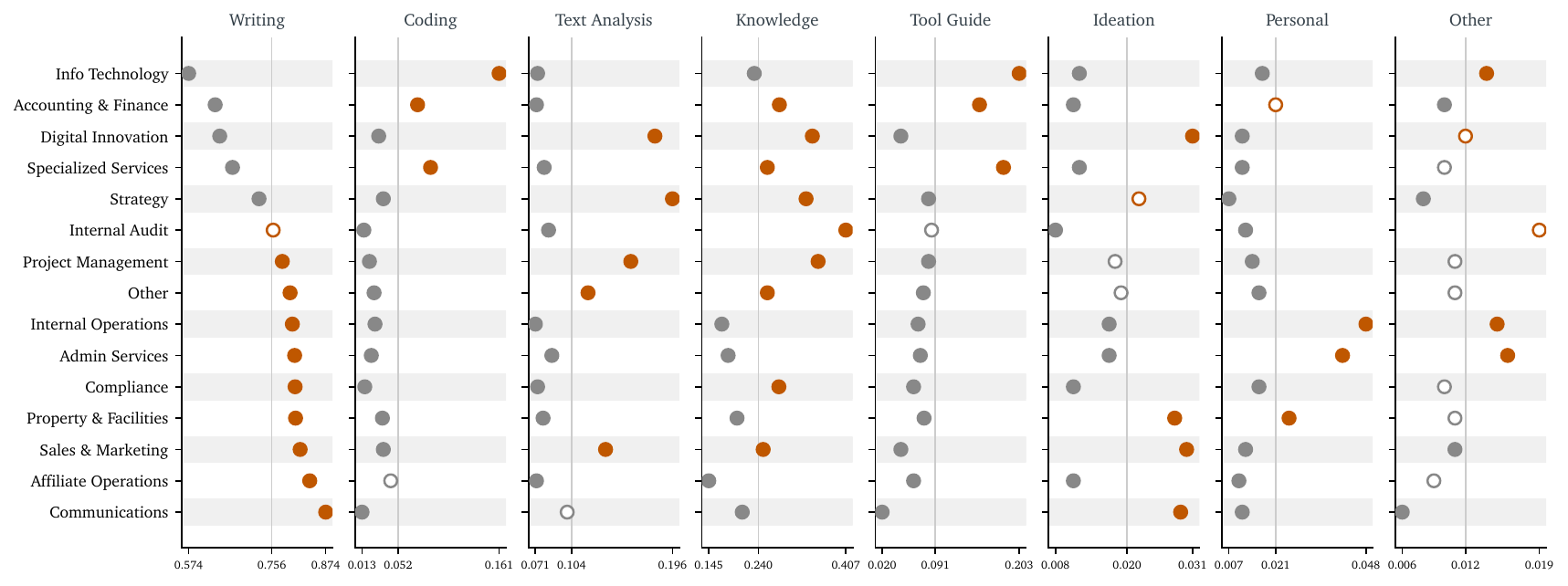}

    \floatfoot{\footnotesize\notesjustified \textit{Notes:} Each panel plots the conversation-level share assigned to each broad use-case category (columns) for the groups listed---months in Panel~A, seniority levels in Panel~B, and functional areas in Panel~C. Each dot marks a group's category share positioned relative to the overall sample mean (the vertical reference line); horizontal-axis ticks report the sample mean at the center and the group minimum and maximum at the ends. Because a conversation may receive multiple labels, proportions do not sum to one. Color and fill follow the same convention as Figure~\ref{fig:adoption}: filled burnt-orange (open burnt-orange) marks values significantly (not significantly) above the sample mean, and filled gray (open gray) marks values significantly (not significantly) below it. Underlying proportions, sample sizes, and two-sided $p$-values are reported in Online Appendix Table~\ref{tab:use_cases_by_group}. Variable definitions are provided in Appendix~\ref{appendix:VariableDefinitions}.}
\end{figure}
\end{landscape}

\clearpage
\begin{figure}[!htbp]
\centering
    \caption{Evaluation of Use}
    \label{fig:evaluation}

    \textit{Panel A: By Month}\par\vspace{0.25em}
    \includegraphics[width=\linewidth]{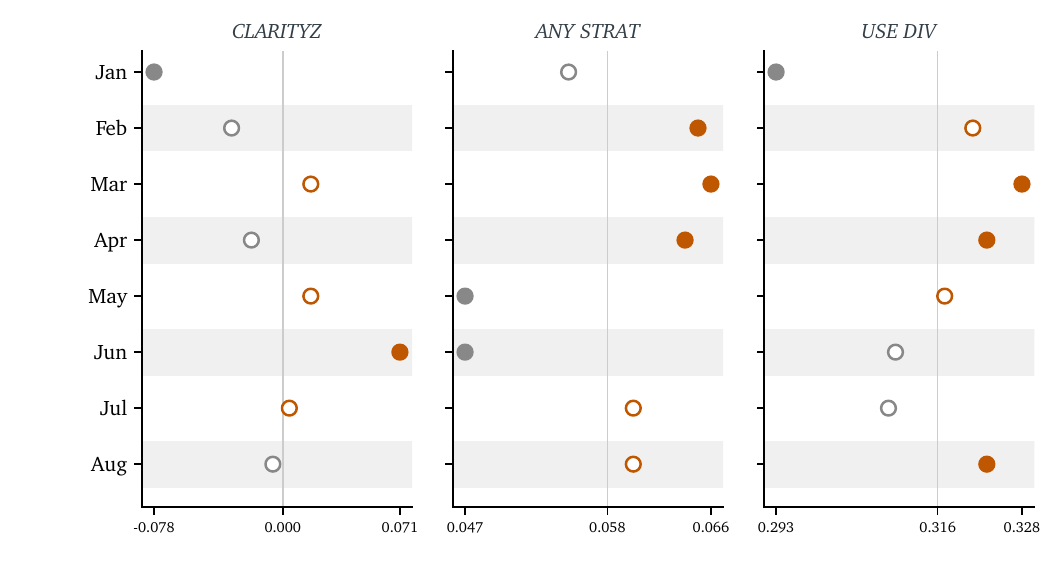}

    \vspace{1.0em}
    \textit{Panel B: By Seniority}\par\vspace{0.25em}
    \includegraphics[width=\linewidth]{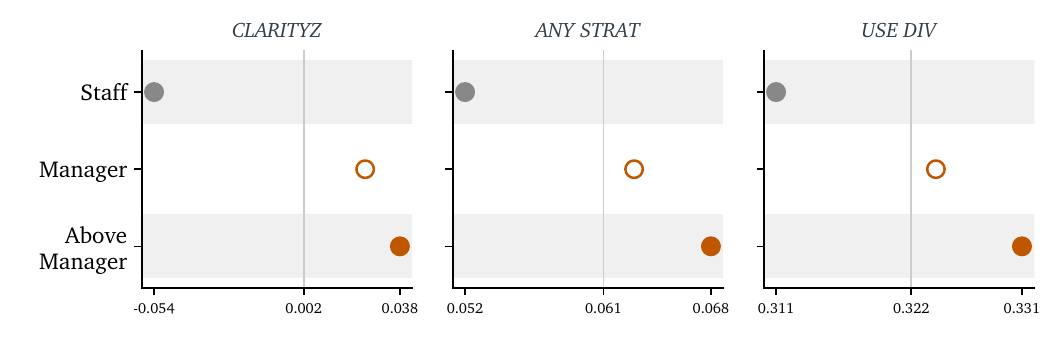}
\end{figure}

\clearpage
\begin{figure}[!htbp]
\ContinuedFloat
\centering
    \caption[]{Evaluation of Use (continued)}

    \textit{Panel C: By Functional Area}\par\vspace{0.25em}
    \includegraphics[width=\linewidth]{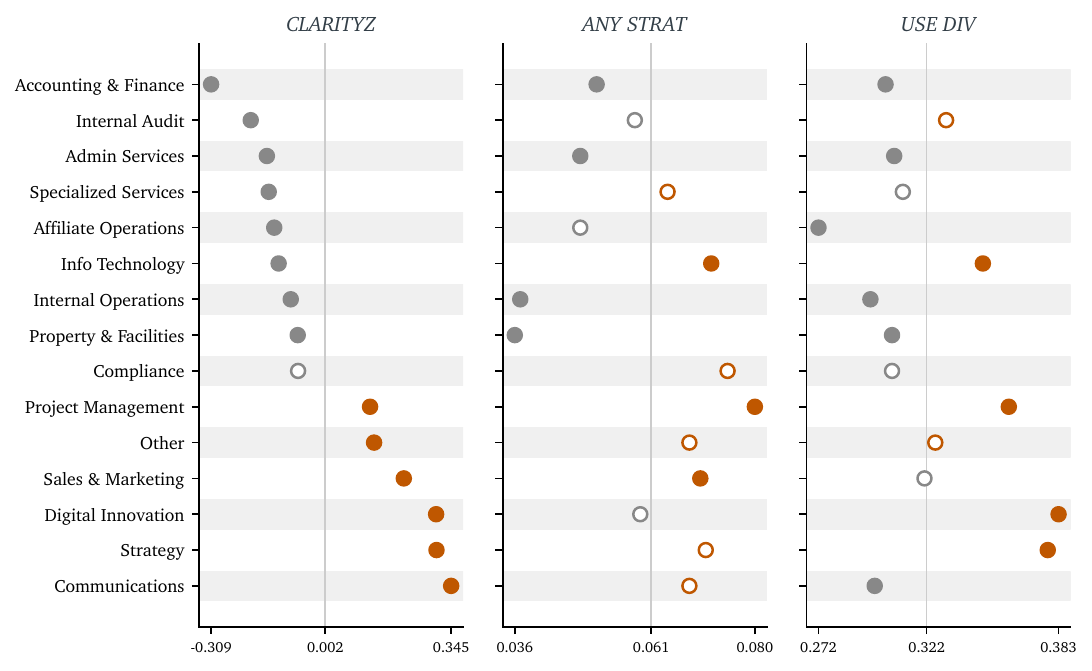}

    \floatfoot{\footnotesize\notesjustified \textit{Notes:} Each panel plots employee-month composite measures of sophisticated use---prompt clarity (\CLARITYZ), deliberate strategy use (\ANYSTRATEGY), and use-case diversity (\USECASEDIVERSITY)---for the groups listed (months in Panel~A, seniority levels in Panel~B, functional areas in Panel~C), computed over active aIQ Chat employee-months. Each dot marks a group mean positioned relative to the overall sample mean (the vertical reference line); horizontal-axis ticks report the sample mean at the center and the group minimum and maximum at the ends. Color and fill follow the same convention as Figure~\ref{fig:adoption}: filled burnt-orange (open burnt-orange) marks values significantly (not significantly) above the sample mean, and filled gray (open gray) marks values significantly (not significantly) below it. Underlying values, sample sizes, and two-sided $p$-values are reported in Online Appendix Table~\ref{tab:evaluation}. Variable definitions are provided in Appendix~\ref{appendix:VariableDefinitions}.}
\end{figure}

\clearpage
\begin{figure}[!htbp]
\centering
    \caption{Bivariate Associations}
    \label{fig:bivar}

    \textit{Panel A: Without User Fixed Effects}\par\vspace{0.25em}
    \includegraphics[width=\linewidth,height=0.66\textheight,keepaspectratio]{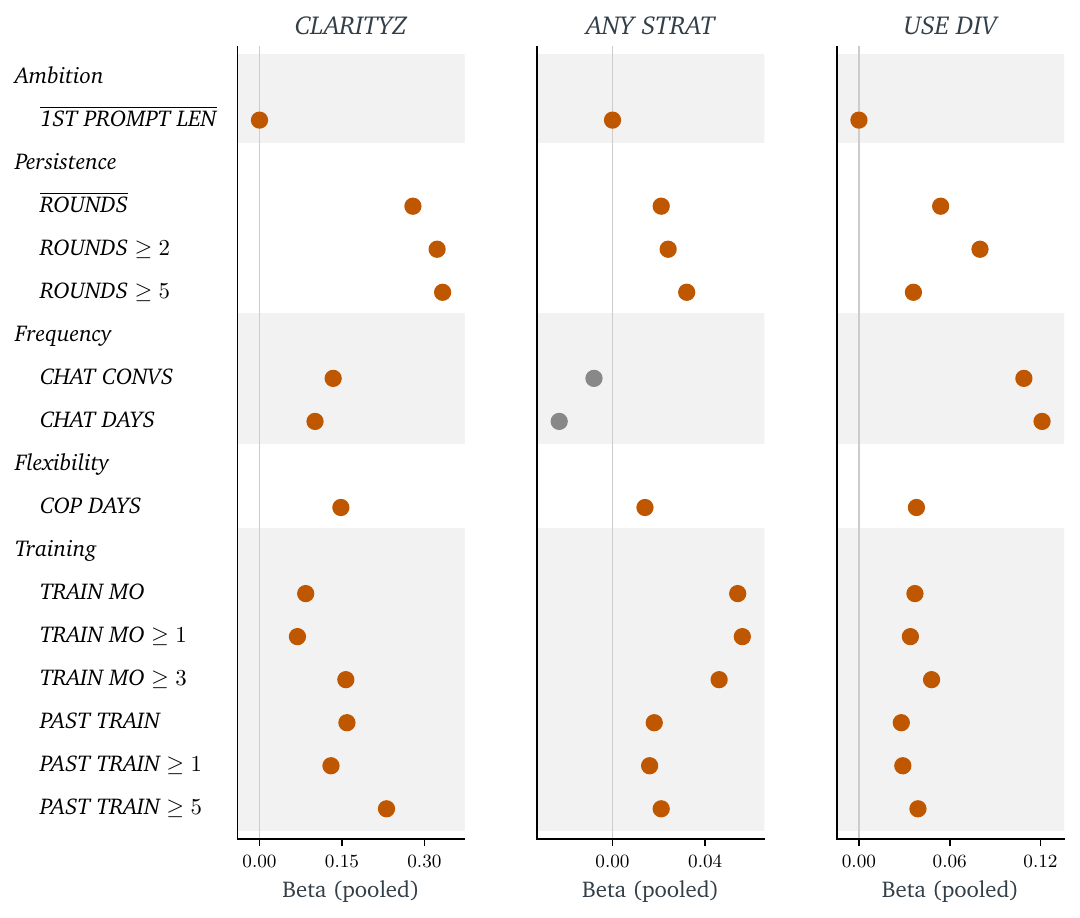}
\end{figure}

\clearpage
\begin{figure}[!htbp]
\ContinuedFloat
\centering
    \caption[]{Bivariate Associations (continued)}

    \textit{Panel B: With User Fixed Effects}\par\vspace{0.25em}
    \includegraphics[width=\linewidth,height=0.66\textheight,keepaspectratio]{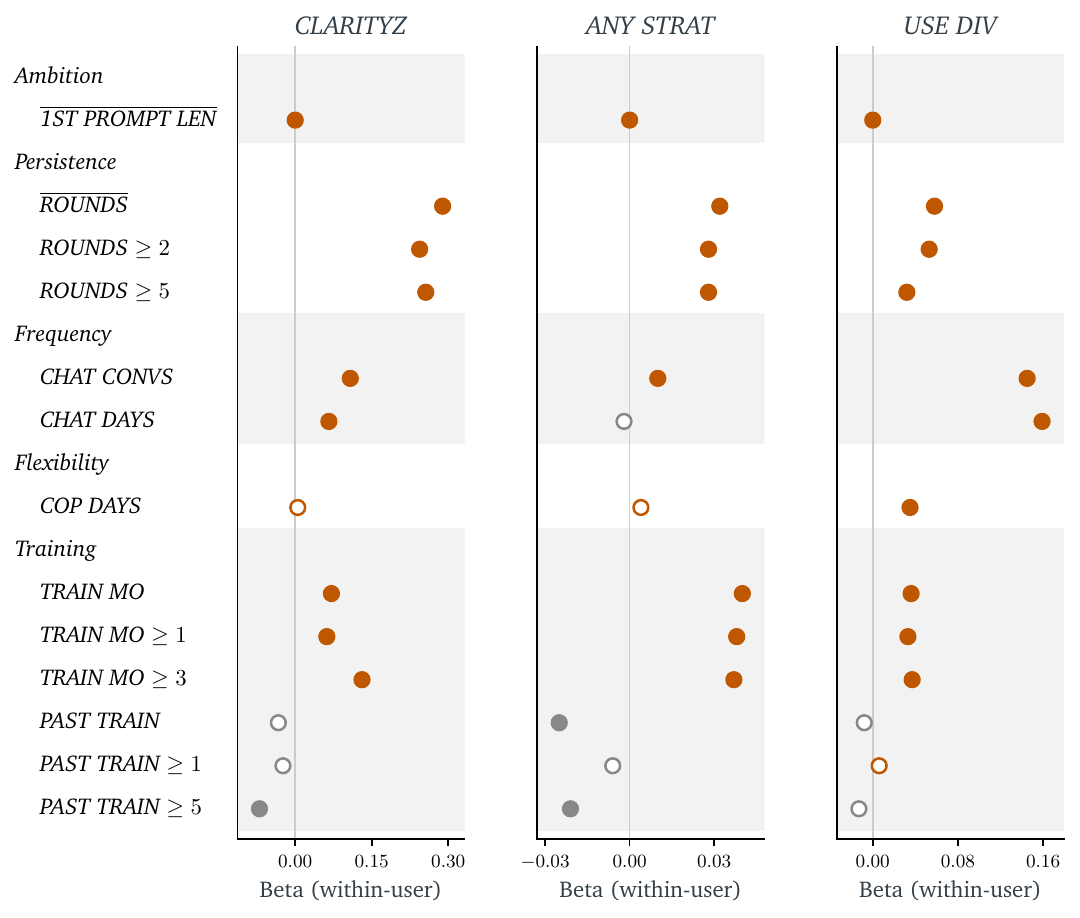}

    \floatfoot{\footnotesize\notesjustified \textit{Notes:} Each panel plots bivariate (pairwise) regression coefficients relating each predictor (rows) to the three sophisticated-use composites: prompt clarity (\CLARITYZ), deliberate strategy use (\ANYSTRATEGY), and use-case diversity (\USECASEDIVERSITY). Panel~A includes month fixed effects only; Panel~B adds user fixed effects. Predictor rows are grouped by construct---ambition, persistence, frequency, flexibility, and training---with alternating blocks shaded, matching Table~\ref{tab:multivar} and Online Appendix Table~\ref{tab:bivar}. Each dot is positioned at the estimated coefficient, with the vertical reference line at zero. Burnt-orange dots denote positive coefficients and gray dots denote negative coefficients; filled dots are statistically significant at the 5\% level (two-sided) and open dots are not. Aside from \UMAVGFIRSTUSERCHARS{} and the threshold indicators, each predictor enters as $\log(1+x)$; row labels omit the transform. For legibility, the figure displays a focused subset of predictors; the complete set of estimates, sample sizes, and $p$-values is reported in Online Appendix Table~\ref{tab:bivar}. Variable definitions are provided in Appendix~\ref{appendix:VariableDefinitions}.}
\end{figure}

\clearpage
\begin{table}[!htbp]
\centering
\begin{threeparttable}[b]
    \caption{Sample Construction and Scale}
    \label{tab:sample_construction}

    \begin{flushleft}
    \textit{Panel A: Employees}\par\vspace{0.25em}
    \input{tables/t1a}

    \vspace{0.75em}
    \textit{Panel B: Conversations}\par\vspace{0.25em}
    \input{tables/t1b}

    \begin{tablenotes}[flushleft]
        \footnotesize
        \item \textit{Notes:} Panel A reports employee counts constructed from employee-month administrative records aggregated to the employee level over the study window. Except for the total count, each Panel A metric requires at least one month of the noted occurrence. Panel B reports descriptive statistics computed from conversation-level logs. The data cover January through August 2025.
    \end{tablenotes}
    \end{flushleft}

\end{threeparttable}
\end{table}

\clearpage
\begin{table}[!htbp]
\centering
\begin{threeparttable}[b]
    \caption{Use Cases}
    \label{tab:use_cases}

    \begin{flushleft}
    \textit{Panel A: Aggregate Categories}\par\vspace{0.25em}
    \input{tables/t3a}

    \vspace{0.75em}
    \textit{Panel B: Subcategories}\par\vspace{0.25em}
    {\footnotesize \input{tables/t3b}\par}

    \begin{tablenotes}[flushleft]
        \footnotesize
        \item \textit{Notes:} Panel A reports the distribution of conversations across broad use-case categories, and Panel B decomposes these categories into subcategories. Conversations may be assigned multiple use-case labels, so category counts and proportions do not sum to one. $N$ is the number of conversations assigned to the category, and ``Pct of Convs'' is the proportion of all conversations in the sample assigned to the category. Variable definitions are provided in Appendix~\ref{appendix:VariableDefinitions}.
    \end{tablenotes}
    \end{flushleft}
\end{threeparttable}
\end{table}

\clearpage
\begin{table}[!htbp]
\centering
\begin{threeparttable}[b]
    \caption{Style and Technique Measures}
    \label{tab:style_technique}
    \begin{flushleft}
    \setlength{\tabcolsep}{4pt}
    \renewcommand{\arraystretch}{0.95}

    \input{tables/t5}

    \begin{tablenotes}[flushleft]
        \footnotesize
        \item \textit{Notes:} The table reports means and standard deviations for prompt input-output measures and strategy measures at the conversation level and for active employee-months. Active employee-months are months with at least one aIQ Chat conversation. The final row in ``Prompt Input-Output Measures'' reports \CLARITYZ, a composite constructed from the measures listed above it. ``Strategy Measures'' are summarized by \ANYSTRATEGY. Although the table also reports conversation-level values for \CLARITYZ{} and \ANYSTRATEGY, all three composite variables are defined at the employee-month level in Appendix~\ref{appendix:VariableDefinitions}. We do not measure \USECASEDIVERSITY{} at the conversation level, as indicated by N/A.
    \end{tablenotes}
    \end{flushleft}
    
\end{threeparttable}
\end{table}

\clearpage
\begin{table}[!htbp]
\centering
\begin{threeparttable}[b]
    \caption{Multivariate Regressions}
    \label{tab:multivar}
    \setlength{\tabcolsep}{3pt}
    \renewcommand{\arraystretch}{0.95}

    \input{tables/t8a}

    \begin{tablenotes}[flushleft]
        \footnotesize
        \item \textit{Notes:} The table reports multivariate associations between the dependent measures and a set of predictors using active employee-month observations with month fixed effects. Columns compare specifications without and with user fixed effects. Coefficients are estimated by OLS after residualizing variables with respect to the included fixed effects, and White robust standard errors are clustered at the user level and reported in parentheses. Statistical significance is indicated by stars appended to coefficients: $^{***}p<0.01$, $^{**}p<0.05$, and $^{*}p<0.10$ (two-sided tests). The $N$ row reports the number of employee-month observations used in each regression. Variable definitions are provided in Appendix~\ref{appendix:VariableDefinitions}.
    \end{tablenotes}
\end{threeparttable}
\end{table}

\clearpage
\singlespacing
\setcounter{section}{0}
\renewcommand{\thesection}{\Alph{section}}

\refstepcounter{section}\label{appendix:VariableDefinitions}
\section*{Appendix \thesection: Variable Definitions}
\small
\input{tables/VariableDefinitions.tex}
\normalsize

\clearpage
\refstepcounter{section}\label{appendix:Tables}
\section*{Online Appendix \thesection: Supplementary Tables}
This appendix is intended for online publication only. It reports the underlying statistics for the figures presented in the main text. Each figure has a corresponding table: Table~\ref{tab:adoption} for Figure~\ref{fig:adoption} (usage and intensity), Table~\ref{tab:use_cases_by_group} for Figure~\ref{fig:use_cases_by_group} (use cases by group), Table~\ref{tab:evaluation} for Figure~\ref{fig:evaluation} (evaluation of use), and Table~\ref{tab:bivar} for Figure~\ref{fig:bivar} (bivariate associations).
\renewcommand{\thetable}{B\arabic{table}}
\setcounter{table}{0}%

\clearpage
\begin{landscape}
\begin{table}[!htbp]
\centering
\begin{threeparttable}[b]
    \caption{Usage and Intensity}
    \label{tab:adoption}
    \setlength{\tabcolsep}{3pt}
    \renewcommand{\arraystretch}{0.95}

    \begin{flushleft}
    \textit{Panel A: By Month}\par\vspace{0.25em}
    \input{tables/t2a}

    \vspace{0.75em}
    \textit{Panel B: By Seniority}\par\vspace{0.25em}
    \input{tables/t2b}
    \end{flushleft}
\end{threeparttable}
\end{table}
\end{landscape}

\clearpage
\begin{landscape}
\begin{table}[!htbp]
\centering
\begin{threeparttable}[b]
    \caption*{\tablename~\thetable: Usage and Intensity (continued)}
    \setlength{\tabcolsep}{3pt}
    \renewcommand{\arraystretch}{0.95}

    \textit{Panel C: By Functional Area}\par\vspace{0.25em}
    \input{tables/t2c}

    \begin{tablenotes}[flushleft]
        \footnotesize
        \item \textit{Notes:} Panels A--C report usage and intensity statistics computed at the employee-month level. Entries are mean-centered by subtracting the sample mean, which is reported in the ``Sample Mean'' row. Statistical significance is assessed group-by-group using a difference-from-rest regression: for each group $g$ and outcome $y$, we estimate $y=\alpha+\beta\cdot\mathbbm{1}\{\text{group}=g\}$ and report two-sided $p$-values for $\beta$ with White robust standard errors. Formatting indicates significance: \textit{italics} $p<0.10$, \textbf{bold} $p<0.05$, and \textbf{\textit{bold italics}} $p<0.01$. Variable definitions are provided in Appendix~\ref{appendix:VariableDefinitions}.
    \end{tablenotes}
\end{threeparttable}
\end{table}
\end{landscape}

\clearpage
\begin{landscape}
\begin{table}[!htbp]
\centering
\begin{threeparttable}[b]
    \caption{Use Cases by Group}
    \label{tab:use_cases_by_group}
    \setlength{\tabcolsep}{3pt}
    \renewcommand{\arraystretch}{0.95}

    \begin{flushleft}
    \textit{Panel A: By Month}\par\vspace{0.25em}
    \input{tables/t4a}

    \vspace{0.75em}
    \textit{Panel B: By Seniority}\par\vspace{0.25em}
    \input{tables/t4b}
    \end{flushleft}
\end{threeparttable}
\end{table}
\end{landscape}

\clearpage
\begin{landscape}
\begin{table}[!htbp]
\centering
\begin{threeparttable}[b]
    \caption*{\tablename~\thetable: Use Cases by Group (continued)}
    \footnotesize
    \setlength{\tabcolsep}{3pt}
    \renewcommand{\arraystretch}{0.95}

    \textit{Panel C: By Functional Area}\par\vspace{0.25em}
    \input{tables/t4c}

    \begin{tablenotes}[flushleft]
        \footnotesize
        \item \textit{Notes:} Panels A--C report the proportion of conversation-level use cases by group. Each cell is the fraction of conversations in the group tagged with the corresponding use-case category. Because conversations may have multiple use-case labels, row proportions may sum to more than one. Panels B and C exclude conversations for which seniority or functional-area data are unavailable. Entries are mean-centered by subtracting the sample mean, which is reported in the ``Sample Mean'' row. Statistical significance is assessed group-by-group using a difference-from-rest regression: for each group $g$ and use-case indicator $y$, we estimate $y=\alpha+\beta\cdot\mathbbm{1}\{\text{group}=g\}$ and report two-sided $p$-values for $\beta$ with White robust standard errors. Formatting indicates significance: \textit{italics} $p<0.10$, \textbf{bold} $p<0.05$, and \textbf{\textit{bold italics}} $p<0.01$. Variable definitions are provided in Appendix~\ref{appendix:VariableDefinitions}.
    \end{tablenotes}
\end{threeparttable}
\end{table}
\end{landscape}

\clearpage
\begin{table}[!htbp]
\centering
\begin{threeparttable}[b]
    \caption{Evaluation of Use}
    \label{tab:evaluation}
    \begin{flushleft}
    \textit{Panel A: By Month}\par\vspace{0.25em}
    \input{tables/t6a}

    \vspace{0.75em}
    \textit{Panel B: By Seniority}\par\vspace{0.25em}
    \input{tables/t6b}
    \end{flushleft}
\end{threeparttable}
\end{table}

\clearpage
\begin{table}[!htbp]
\centering
\begin{threeparttable}[b]
    \caption*{\tablename~\thetable: Evaluation of Use (continued)}
    \setlength{\tabcolsep}{3pt}
    \renewcommand{\arraystretch}{0.95}

    \begin{flushleft}
    \textit{Panel C: By Functional Area}\par\vspace{0.25em}
    \input{tables/t6c}

    \begin{tablenotes}[flushleft]
        \footnotesize
        \item \textit{Notes:} Panels A--C report group averages computed over active employee-months. Active employee-months are months with at least one aIQ Chat conversation. $N$ is the number of active employee-months in the group. Panels B and C exclude active employee-months for which seniority or functional-area data are unavailable. Entries are mean-centered by subtracting the sample mean, which is reported in the ``Sample Mean'' row; percentage-formatted columns are shown as percentage-point deviations from the sample mean. Statistical significance is assessed group-by-group using a difference-from-rest regression: for each group $g$ and outcome $y$, we estimate $y=\alpha+\beta\cdot\mathbbm{1}\{\text{group}=g\}$ and report two-sided $p$-values for $\beta$ with White robust standard errors. Formatting indicates significance: \textit{italics} $p<0.10$, \textbf{bold} $p<0.05$, and \textbf{\textit{bold italics}} $p<0.01$. Variable definitions are provided in Appendix~\ref{appendix:VariableDefinitions}.
    \end{tablenotes}
    \end{flushleft}
\end{threeparttable}
\end{table}

\clearpage
\begin{table}[!htbp]
\centering
\begin{threeparttable}[b]
    \caption{Bivariate Associations}
    \label{tab:bivar}
    \setlength{\tabcolsep}{3pt}
    \renewcommand{\arraystretch}{0.95}

    \textit{Panel A: Without User Fixed Effects}\par\vspace{0.25em}
    \input{tables/t7a}
\end{threeparttable}
\end{table}

\clearpage
\begin{table}[!htbp]
\centering
\begin{threeparttable}[b]
    \caption*{\tablename~\thetable: Bivariate Associations (continued)}
    \setlength{\tabcolsep}{3pt}
    \renewcommand{\arraystretch}{0.95}

    \textit{Panel B: With User Fixed Effects}\par\vspace{0.25em}
    \input{tables/t7b}

    \begin{tablenotes}[flushleft]
        \footnotesize
        \item \textit{Notes:} Panels A and B report bivariate associations between each dependent measure and each predictor using active employee-month observations with month fixed effects; Panel B additionally includes user fixed effects. Predictors are grouped by the construct each one proxies for, following Figure~\ref{fig:bivar}. Aside from \UMAVGFIRSTUSERCHARS{} and the threshold indicators, each predictor enters as $\log(1+x)$; row labels omit the transform for legibility. Coefficients are estimated by OLS after residualizing variables with respect to the included fixed effects. White robust standard errors are clustered at the user level. The table reports coefficient estimates and corresponding two-sided $p$-values. The $N$ column reports the number of active employee-months with non-missing predictor values; the effective regression sample may vary across dependent measures because of predictor missingness. Variable definitions are provided in Appendix~\ref{appendix:VariableDefinitions}.
    \end{tablenotes}
\end{threeparttable}
\end{table}

\clearpage
\refstepcounter{section}\label{appendix:MetaPrompts}
\section*{Online Appendix \thesection: Meta Prompts}
This appendix is intended for online publication only. It reproduces the metaprompts submitted with each aIQ Chat transcript.
\input{tables/MetaPrompts.tex}

\end{document}

%% file: tables/t1a.tex
\begin{tabular}{lr}
\toprule
Metric & Count \\
\midrule
Total back-office employees & 5,191 \\
Employees with aIQ Chat use & 3,925 \\
Employees in Copilot dataset & 4,392 \\
Employees with Copilot use & 3,962 \\
Employees who used BOTH tools & 2,861 \\
\bottomrule
\end{tabular}

%% file: tables/t1b.tex
\begin{tabular}{lr}
\toprule
Metric & Value \\
\midrule
Number of conversations & 158,496 \\
Number of submitted user prompts & 713,564 \\
Mean user prompts per conversation & 4.50 \\
Median user prompts per conversation & 2.00 \\
Number of active employees (with LLM use) & 3,925 \\
Mean conversations per active employee & 40.38 \\
Median conversations per active employee & 14.00 \\
Share of conversations by top decile of users & 0.51 \\
\bottomrule
\end{tabular}

%% file: tables/t3a.tex
\begin{tabular}{lrr}
\toprule
Category & N & Pct of Convs \\
\midrule
\WRITINGCOMMUNICATION & 116,216 & 0.733 \\
\KNOWLEDGEEXPERTISE & 36,458 & 0.230 \\
\TEXTBASEDANALYSIS & 17,008 & 0.107 \\
\SOFTWARETOOLGUIDANCE & 15,634 & 0.099 \\
\CODINGDATAANALYSIS & 11,556 & 0.073 \\
\PERSONALNONWORK & 3,026 & 0.019 \\
\CREATIVETHINKINGIDEATION & 2,910 & 0.018 \\
\OTHER & 1,930 & 0.012 \\
\bottomrule
\end{tabular}

%% file: tables/t3b.tex
\begin{tabular}{llrr}
\toprule
Category & Subcategory & N & Pct of Convs \\
\midrule
\WRITINGCOMMUNICATION & Edit/improve existing content & 72,596 & 0.458 \\
 & Generate new content & 55,125 & 0.348 \\
 & Language translation & 1,549 & 0.010 \\
 & Other writing tasks & 885 & 0.006 \\
\CODINGDATAANALYSIS & Generate or debug programming code & 8,546 & 0.054 \\
 & Analyze user-provided data & 1,520 & 0.010 \\
 & Data cleaning / restructuring / standardization & 1,046 & 0.007 \\
 & Other coding/data tasks & 628 & 0.004 \\
\TEXTBASEDANALYSIS & Document understanding (text-based analysis) & 16,524 & 0.104 \\
 & Other text-based analysis requests & 488 & 0.003 \\
\KNOWLEDGEEXPERTISE & Business management (back office) & 14,432 & 0.091 \\
 & Other knowledge queries & 9,777 & 0.062 \\
 & Business operations (front office) & 8,847 & 0.056 \\
 & Legal / regulation / compliance & 4,050 & 0.026 \\
 & Accounting standards and guidance & 1,255 & 0.008 \\
\SOFTWARETOOLGUIDANCE & Other software (Sheets, Tableau, etc.) & 9,812 & 0.062 \\
 & Microsoft Excel & 3,540 & 0.022 \\
 & AI/LLM tools & 1,037 & 0.007 \\
 & Microsoft Outlook & 802 & 0.005 \\
 & Microsoft PowerPoint & 531 & 0.003 \\
 & Alteryx & 428 & 0.003 \\
 & Microsoft Word & 391 & 0.002 \\
\CREATIVETHINKINGIDEATION & Brainstorming and ideation & 2,822 & 0.018 \\
 & Other creative thinking requests & 88 & 0.001 \\
\PERSONALNONWORK & Personal tasks & 2,075 & 0.013 \\
 & Other non-work activities & 959 & 0.006 \\
\OTHER & Other / fits nowhere else & 1,315 & 0.008 \\
 & Unclear or ambiguous use case & 474 & 0.003 \\
 & Testing LLM capabilities & 141 & 0.001 \\
\bottomrule
\end{tabular}

%% file: tables/t5.tex
\begin{tabular}{lrrrr}
\toprule
 & \multicolumn{2}{r}{Conversations} & \multicolumn{2}{r}{Employee-Months} \\
 & \multicolumn{2}{r}{N = 158,496} & \multicolumn{2}{r}{N = 17,671} \\
Variable & Mean & Std. Dev. & Mean & Std. Dev. \\
\midrule
\textit{Prompt Input-Output Measures:} &  &  &  &  \\
\INITIALLANGUAGECOMPLEXITY & 2.586 & 0.724 & 2.474 & 0.442 \\
\INITIALCOMPLEXITY & 2.305 & 0.665 & 2.272 & 0.372 \\
\PROMPTSTRUCTUREQUALITY & 1.738 & 0.734 & 1.629 & 0.404 \\
\USERPROMPTINGSOPHISTICATION & 2.294 & 0.693 & 2.229 & 0.409 \\
\SPECIFICITYSCORE & 3.251 & 0.689 & 3.187 & 0.426 \\
\OUTPUTFORMATCLARITY & 1.812 & 0.841 & 1.753 & 0.477 \\
\CONSTRAINTSPRESENT & 0.243 & 0.429 & 0.238 & 0.299 \\
\STRUCTUREDOUTPUTREQUESTED & 0.029 & 0.167 & 0.027 & 0.117 \\
\ACCEPTANCECRITERIAPROVIDED & 0.010 & 0.100 & 0.006 & 0.055 \\
\CLARITYZ & 0.000 & 1.000 & 0.000 & 1.000 \\
 &  &  &  &  \\
\textit{Strategy Measures} &  &  &  &  \\
\ROLEPLAY & 0.044 & 0.205 & 0.050 & 0.171 \\
\FEWSHOTEXAMPLES & 0.003 & 0.057 & 0.003 & 0.035 \\
\CHAINOFTHOUGHT & 0.001 & 0.038 & 0.001 & 0.021 \\
\SELFVERIFICATION & 0.003 & 0.054 & 0.002 & 0.028 \\
\INTERACTIVEREFINEMENT & 0.005 & 0.070 & 0.010 & 0.074 \\
\ANYSTRATEGY & 0.052 & 0.221 & 0.058 & 0.183 \\
 &  &  &  &  \\
\textit{Other Prompt Measures} &  &  &  &  \\
\USECASEDIVERSITY & \multicolumn{2}{c}{N/A} & 0.316 & 0.222 \\
\DOCUMENTUPLOADED & 0.055 & 0.227 & 0.058 & 0.165 \\
\USERSATISFIED & 0.035 & 0.184 & 0.011 & 0.081 \\
\FREQUENTTYPOS & 0.107 & 0.308 & 0.089 & 0.195 \\
\PLEASANTRIESUSED & 0.302 & 0.459 & 0.093 & 0.226 \\
\ALLLOWERCASE & 0.054 & 0.227 & 0.073 & 0.187 \\
\bottomrule
\end{tabular}

%% file: tables/t8a.tex
\begin{tabular}{lrrrrrr}
\toprule
Variable & \CLARITYZ & \CLARITYZ & \ANYSTRATEGY & \ANYSTRATEGY & \USECASEDIVERSITY & \USECASEDIVERSITY \\
\midrule
\UMAVGFIRSTUSERCHARS & 0.000*** & 0.000*** & 0.000*** & 0.000**\phantom{*} & 0.000*** & 0.000*** \\
 & (0.000) & (0.000) & (0.000) & (0.000) & (0.000) & (0.000) \\
\LOGUMAVGUSERROUNDS & 0.259*** & 0.257*** & 0.042*** & 0.044*** & 0.014*** & 0.023*** \\
 & (0.020) & (0.022) & (0.005) & (0.005) & (0.005) & (0.005) \\
\addlinespace
\LOGGPTUSAGEDATES & 0.005\phantom{***} & 0.015\phantom{***} & -0.043*** & -0.018*** & 0.103*** & 0.144*** \\
 & (0.017) & (0.019) & (0.004) & (0.004) & (0.004) & (0.005) \\
\LOGCOPILOTUSAGEDATES & 0.106*** & -0.016\phantom{***} & 0.013*** & 0.003\phantom{***} & 0.019*** & 0.005\phantom{***} \\
 & (0.012) & (0.016) & (0.002) & (0.003) & (0.003) & (0.004) \\
\addlinespace
\LOGTRAININGSCOMPLETEDMONTH & 0.046**\phantom{*} & 0.051*** & 0.049*** & 0.037*** & 0.034*** & 0.028*** \\
 & (0.019) & (0.019) & (0.004) & (0.004) & (0.004) & (0.004) \\
\LOGTRAININGSCOMPLETEDSTARTMONTH & 0.099*** & 0.002\phantom{***} & 0.013*** & -0.007\phantom{***} & 0.016*** & 0.003\phantom{***} \\
 & (0.020) & (0.025) & (0.004) & (0.006) & (0.004) & (0.006) \\
\addlinespace
\UMALLLOWERCASERATE & -1.310*** & -1.084*** & -0.080*** & -0.064*** & -0.117*** & -0.116*** \\
 & (0.050) & (0.067) & (0.007) & (0.014) & (0.010) & (0.014) \\
\UMFREQUENTTYPOSRATE & -0.117\phantom{***} & -0.007\phantom{***} & -0.069*** & -0.042**\phantom{*} & -0.050*** & -0.047*** \\
 & (0.086) & (0.098) & (0.019) & (0.019) & (0.016) & (0.017) \\
\UMDOCUMENTUPLOADEDRATE & 0.653*** & 0.744*** & 0.009\phantom{***} & -0.004\phantom{***} & 0.042**\phantom{*} & 0.045**\phantom{*} \\
 & (0.106) & (0.118) & (0.023) & (0.023) & (0.018) & (0.021) \\
\PLEASANTRIESUSEDRATE & 0.398*** & 0.363*** & -0.011\phantom{***} & -0.013\phantom{***} & -0.116*** & -0.083*** \\
 & (0.055) & (0.058) & (0.013) & (0.012) & (0.009) & (0.011) \\
\USERSATISFIEDRATE & 0.064\phantom{***} & -0.179\phantom{***} & 0.013\phantom{***} & -0.035\phantom{***} & 0.016\phantom{***} & 0.013\phantom{***} \\
 & (0.134) & (0.145) & (0.036) & (0.040) & (0.026) & (0.030) \\
\midrule
Month FE & Yes & Yes & Yes & Yes & Yes & Yes \\
User FE & No & Yes & No & Yes & No & Yes \\
N & 12,902 & 12,902 & 12,902 & 12,902 & 12,902 & 12,902 \\
\bottomrule
\end{tabular}

%% file: tables/VariableDefinitions.tex
\begin{longtable}{>{\raggedright\arraybackslash}p{3.1cm}p{\dimexpr\textwidth-3.1cm-4\tabcolsep\relax}}

    \\[-1.8ex]\hline \hline \\[-1.8ex]
    \endfirsthead

    \multicolumn{2}{l}%
    {{\textit{Continued from previous page}}} \\
    \midrule
    \endhead

    \multicolumn{2}{r}{{\textit{Continued on next page}}} 
    \endfoot

    \multicolumn{2}{r}{{ }} 
    \endlastfoot

    \\ \multicolumn{2}{l}{\textbf{Sophisticated use measures}}\\*
    \midrule

    \CLARITYZ & Composite (z-scored) measure of prompt clarity for employee $i$ in month $t$. Constructed by first z-scoring, within the sample of active employee-months, the employee-month averages of six LLM-assigned continuous prompt-quality ratings (\INITIALLANGUAGECOMPLEXITY, \INITIALCOMPLEXITY, \PROMPTSTRUCTUREQUALITY, \USERPROMPTINGSOPHISTICATION, \SPECIFICITYSCORE, and \OUTPUTFORMATCLARITY) and three LLM-assigned binary prompt features (\CONSTRAINTSPRESENT, \STRUCTUREDOUTPUTREQUESTED, and \ACCEPTANCECRITERIAPROVIDED). The composite is the equally weighted average of these nine z-scores and is then z-scored again within active employee-months. Higher values indicate clearer, more structured prompts.\\
    \midrule

    \ANYSTRATEGY & Composite measure of deliberate prompting strategy use for employee $i$ in month $t$. Constructed as the proportion of the employee-month's conversations in which at least one of five prompting strategies equals one (\ROLEPLAY, \FEWSHOTEXAMPLES, \CHAINOFTHOUGHT, \SELFVERIFICATION, or \INTERACTIVEREFINEMENT).\\
    \midrule

    \USECASEDIVERSITY & 
    Measure of how broadly an employee’s LLM conversations span different use-case categories in a given month. We compute it as a normalized Shannon-entropy index across that employee-month’s totals for the eight broad use-case categories listed below, counting each conversation once in every broad category to which it is assigned: for each category, we convert its monthly total into a share of the employee’s total across all categories, then compute entropy and normalize it to range from 0 to 1. The measure equals 0 when all activity falls in a single category (or when the employee-month has a total of zero across categories) and approaches 1 when activity is evenly spread across categories. \\
    \midrule

    \\ \multicolumn{2}{l}{\textbf{Use-case categories}}\\*
    \midrule

    \WRITINGCOMMUNICATION & Indicator equal to 1 if the conversation includes writing or communication tasks (e.g., generating new content, editing user-provided text, translation, or other writing tasks), and 0 otherwise.\\
    \midrule

    \CODINGDATAANALYSIS & Indicator equal to 1 if the conversation includes coding or data analysis tasks (e.g., generating/debugging code, analyzing user-provided data, or cleaning/restructuring data), and 0 otherwise.\\
    \midrule

    \TEXTBASEDANALYSIS & Indicator equal to 1 if the conversation includes text-based analysis or document understanding (e.g., summarizing or interpreting user-provided text/documents), and 0 otherwise.\\
    \midrule

    \KNOWLEDGEEXPERTISE & Indicator equal to 1 if the conversation seeks domain knowledge or expertise (e.g., accounting, legal/regulatory, business operations/management, or other knowledge queries), and 0 otherwise.\\
    \midrule

    \SOFTWARETOOLGUIDANCE & Indicator equal to 1 if the conversation requests guidance on how to use a specific software/tool (e.g., Excel, Word, PowerPoint, Outlook, AI/LLM tools), and 0 otherwise.\\
    \midrule

    \CREATIVETHINKINGIDEATION & Indicator equal to 1 if the conversation involves brainstorming or ideation (generating new ideas), and 0 otherwise.\\
    \midrule

    \PERSONALNONWORK & Indicator equal to 1 if the conversation is primarily non-work/personal (e.g., personal planning, home projects), and 0 otherwise.\\
    \midrule

    \OTHER & Indicator equal to 1 if the conversation is classified as Other in the use-case taxonomy (including testing LLM capabilities or unclear/ambiguous use cases), and 0 otherwise.\\
    \midrule

    \\ \multicolumn{2}{l}{\textbf{Predictor variables (described at employee-month level)}}\\*
    \midrule

    \UMAVGFIRSTUSERCHARS & Employee-month average number of characters in the first substantive user prompt of each conversation.\\
    \midrule

    \LOGUMAVGUSERROUNDS & Natural logarithm of $1+$ the average number of user turns per conversation for employee $i$ in month $t$.\\
    \midrule

    \LOGUMNUMCONVERSATIONS & Natural logarithm of $1+$ the number of aIQ Chat conversations for employee $i$ in month $t$.\\
    \midrule

    \LOGGPTUSAGEDATES & Natural logarithm of $1+$ the number of days in month $t$ with any aIQ Chat usage.\\
    \midrule

    \LOGCOPILOTUSAGEDATES & Natural logarithm of $1+$ the number of days in month $t$ with any Copilot usage.\\
    \midrule

    \LOGTRAININGSCOMPLETEDMONTH & Natural logarithm of $1+$ the number of training courses completed by employee $i$ in month $t$.\\
    \midrule

    \LOGTRAININGSCOMPLETEDSTARTMONTH & Natural logarithm of $1+$ the cumulative number of training courses completed by employee $i$ through the beginning of month $t$.\\
    \midrule

    \\ \multicolumn{2}{l}{\textbf{Threshold indicators (described at employee-month level)}}\\*
    \midrule

    \USERROUNDSGE{k} & Indicator equal to 1 if the employee-month average number of user turns per conversation is at least $k$, and 0 otherwise, for $k\in\{2,3,5\}$.\\
    \midrule

    \GPTDAYSGE{k} & Indicator equal to 1 if aIQ Chat usage days in month $t$ are at least $k$, and 0 otherwise, for $k\in\{5,10,15\}$.\\
    \midrule

    \COPILOTDAYSGE{k} & Indicator equal to 1 if Copilot usage days in month $t$ are at least $k$, and 0 otherwise, for $k\in\{1,5,10,15\}$.\\
    \midrule

    \TRAININGSMONTHGE{k} & Indicator equal to 1 if training courses completed by employee $i$ in month $t$ are at least $k$, and 0 otherwise, for $k\in\{1,3,5\}$.\\
    \midrule

    \TRAININGSSTARTGE{k} & Indicator equal to 1 if cumulative training courses completed by employee $i$ through month $t$ are at least $k$, and 0 otherwise, for $k\in\{1,3,5,10\}$.\\
    \midrule

    \\ \multicolumn{2}{l}{\textbf{Prompt input-output measures}}\\*
    \midrule

    \INITIALLANGUAGECOMPLEXITY & LLM-assigned rating of linguistic complexity in the first substantive user prompt (1 = very simple language; 5 = very complex language).\\
    \midrule

    \INITIALCOMPLEXITY & LLM-assigned rating of structural/task complexity in the first substantive user prompt (1 = trivial request; 5 = very complex request).\\
    \midrule

    \PROMPTSTRUCTUREQUALITY & LLM-assigned rating of prompt organization and structure (1 = unstructured; 5 = highly structured).\\
    \midrule

    \USERPROMPTINGSOPHISTICATION & LLM-assigned rating of prompting sophistication (strategy-agnostic) across the conversation (1 = naive; 5 = expert).\\
    \midrule

    \SPECIFICITYSCORE & LLM-assigned rating of overall specificity and precision across the conversation (1 = very vague; 5 = very specific).\\
    \midrule

    \OUTPUTFORMATCLARITY & LLM-assigned rating of how clearly the user specifies the desired output format (1 = no output guidance; 5 = precise specification).\\
    \midrule

    \CONSTRAINTSPRESENT & Indicator equal to 1 if the initial substantive prompt includes at least one explicit, enforceable constraint (e.g., length, schema, tone, audience, timeframe), and 0 otherwise.\\
    \midrule

    \STRUCTUREDOUTPUTREQUESTED & Indicator equal to 1 if the user explicitly requests structured data output (e.g., JSON, YAML, CSV, or a table with specified fields/schema), and 0 otherwise.\\
    \midrule

    \ACCEPTANCECRITERIAPROVIDED & Indicator equal to 1 if the user provides explicit acceptance criteria or success tests (e.g., must include specific fields or pass explicit checks), and 0 otherwise.\\
    \midrule

    \\ \multicolumn{2}{l}{\textbf{Prompting strategies}}\\*
    \midrule

    \ROLEPLAY & Indicator equal to 1 if the user assigns the LLM a specific role/persona (e.g., ``act as an auditor''), and 0 otherwise.\\
    \midrule

    \FEWSHOTEXAMPLES & Indicator equal to 1 if the user provides concrete input--output examples to guide the LLM, and 0 otherwise.\\
    \midrule

    \CHAINOFTHOUGHT & Indicator equal to 1 if the user explicitly requests step-by-step reasoning or asks to see the LLM's reasoning process, and 0 otherwise.\\
    \midrule

    \SELFVERIFICATION & Indicator equal to 1 if the user instructs the LLM to check, verify, or validate its own output, and 0 otherwise.\\
    \midrule

    \INTERACTIVEREFINEMENT & Indicator equal to 1 if the user requests collaborative back-and-forth to refine the task (e.g., asks the LLM to ask clarifying questions), and 0 otherwise.\\
    \midrule

    \\ \multicolumn{2}{l}{\textbf{Style rates (described at employee-month level)}}\\*
    \midrule

    \UMALLLOWERCASERATE & Share of employee $i$'s conversations in month $t$ in which all user messages are entirely lowercase (excluding quoted text, URLs, file paths, code snippets, and other exempt content).\\
    \midrule

    \UMFREQUENTTYPOSRATE & Share of employee $i$'s conversations in month $t$ in which the user makes frequent typos (at least 3 typos across the conversation or at least 2 typos in a single message).\\
    \midrule

    \UMDOCUMENTUPLOADEDRATE & Share of employee $i$'s conversations in month $t$ in which the user uploads or references an attached document.\\
    \midrule

    \PLEASANTRIESUSED & Share of employee $i$'s conversations in month $t$ in which the user includes politeness expressions (e.g., greetings, thanks, ``please'').\\
    \midrule

    \USERSATISFIEDRATE & Share of employee $i$'s conversations in month $t$ in which the user expresses explicit positive feedback (e.g., ``thanks'' or ``perfect'').\\
    \midrule

    \\[-1.8ex]\hline \hline \\[-1.8ex]

\end{longtable}

%% file: tables/t2a.tex
\begin{tabular}{lrrrrrrr}
\toprule
Month & N & \GPTORCOPILOT & \GPTUSE & \COPILOTUSE & \GPTMEANDAYS & \COPILOTMEANDAYS & \GPTCONVSMEAN \\
\midrule
Sample Mean & 32,849 & 0.839 & 0.437 & 0.780 & 2.714 & 5.832 & 8.563 \\
2025-01 & 3,928 & \textbf{\textit{-0.060}} & \textbf{\textit{-0.051}} & \textbf{\textit{-0.070}} & \textbf{\textit{-0.464}} & \textbf{\textit{-1.360}} & -0.117 \\
2025-02 & 4,101 & \textbf{\textit{-0.034}} & -0.016 & \textbf{\textit{-0.042}} & \textbf{\textit{-0.191}} & \textbf{\textit{-0.611}} & 0.342 \\
2025-03 & 4,105 & \textbf{0.006} & 0.008 & 0.003 & \textbf{\textit{0.193}} & \textbf{-0.169} & \textbf{0.506} \\
2025-04 & 4,078 & \textbf{0.020} & 0.012 & \textbf{\textit{0.023}} & \textbf{0.180} & \textbf{\textit{0.323}} & \textit{0.480} \\
2025-05 & 4,064 & \textbf{\textit{0.033}} & \textbf{0.020} & \textbf{\textit{0.038}} & \textbf{\textit{0.328}} & \textbf{\textit{0.583}} & 0.135 \\
2025-06 & 4,235 & 0.009 & 0.019 & 0.006 & 0.006 & -0.102 & -0.062 \\
2025-07 & 4,125 & \textbf{\textit{-0.014}} & \textbf{-0.009} & \textbf{\textit{-0.011}} & \textit{-0.125} & -0.015 & \textbf{\textit{-0.957}} \\
2025-08 & 4,213 & \textbf{\textit{0.036}} & 0.013 & \textbf{\textit{0.048}} & 0.058 & \textbf{\textit{1.269}} & -0.346 \\
\bottomrule
\end{tabular}

%% file: tables/t2b.tex
\begin{tabular}{lrrrrrrr}
\toprule
Seniority & N & \GPTORCOPILOT & \GPTUSE & \COPILOTUSE & \GPTMEANDAYS & \COPILOTMEANDAYS & \GPTCONVSMEAN \\
\midrule
Sample Mean & 32,849 & 0.839 & 0.437 & 0.780 & 2.714 & 5.832 & 8.563 \\
Staff & 12,360 & \textbf{\textit{-0.050}} & \textbf{\textit{-0.019}} & \textbf{\textit{-0.064}} & \textbf{-0.075} & \textbf{\textit{-0.821}} & \textbf{\textit{0.416}} \\
Manager & 7,398 & -0.002 & \textbf{\textit{0.024}} & \textbf{-0.010} & \textbf{\textit{0.356}} & \textbf{0.122} & \textbf{\textit{0.450}} \\
Above Manager & 13,091 & \textbf{\textit{0.048}} & 0.004 & \textbf{\textit{0.066}} & \textbf{\textit{-0.130}} & \textbf{\textit{0.706}} & \textbf{\textit{-0.639}} \\
\bottomrule
\end{tabular}

%% file: tables/t2c.tex
\begin{tabular}{lrrrrrrr}
\toprule
\makecell[l]{Functional\\Area} & N & \GPTORCOPILOT & \GPTUSE & \COPILOTUSE & \GPTMEANDAYS & \COPILOTMEANDAYS & \GPTCONVSMEAN \\
\midrule
Sample Mean & 32,849 & 0.839 & 0.437 & 0.780 & 2.714 & 5.832 & 8.563 \\
Accounting \& Finance & 2,967 & \textbf{\textit{-0.057}} & \textbf{\textit{-0.039}} & \textbf{\textit{-0.043}} & \textbf{\textit{-0.447}} & \textbf{\textit{-0.768}} & \textbf{\textit{-1.637}} \\
Administrative Services & 2,995 & 0.004 & \textbf{\textit{0.031}} & 0.005 & 0.048 & \textbf{-0.215} & 0.118 \\
Affiliate Operations & 657 & \textbf{-0.033} & \textbf{\textit{-0.122}} & -0.001 & \textbf{\textit{-0.877}} & \textbf{\textit{-0.981}} & -0.776 \\
Communications & 827 & \textbf{\textit{0.088}} & \textbf{\textit{0.115}} & \textbf{\textit{0.110}} & \textbf{\textit{0.847}} & \textbf{\textit{2.034}} & 0.481 \\
Compliance & 865 & \textbf{\textit{-0.105}} & \textbf{\textit{-0.092}} & \textbf{\textit{-0.189}} & \textbf{-0.362} & \textbf{\textit{-1.308}} & \textit{1.343} \\
Digital Innovation & 287 & \textbf{\textit{0.098}} & \textbf{\textit{0.218}} & \textbf{\textit{0.115}} & \textbf{\textit{0.728}} & \textbf{\textit{1.464}} & \textbf{\textit{-1.553}} \\
Information Technology & 6,931 & \textbf{\textit{-0.037}} & \textbf{\textit{-0.075}} & \textbf{\textit{-0.040}} & \textbf{\textit{-0.640}} & 0.070 & \textbf{\textit{-1.294}} \\
Internal Audit & 158 & \textbf{\textit{0.104}} & \textbf{\textit{0.196}} & \textbf{\textit{0.131}} & \textbf{\textit{2.254}} & 0.345 & -0.063 \\
Internal Operations & 2,560 & \textbf{\textit{0.042}} & \textbf{\textit{0.070}} & \textbf{\textit{0.039}} & \textbf{\textit{0.348}} & \textbf{0.269} & 0.206 \\
Other & 4,663 & 0.005 & \textbf{\textit{-0.101}} & \textbf{\textit{0.027}} & \textbf{\textit{-0.520}} & 0.030 & \textbf{\textit{1.070}} \\
Project Management & 1,036 & \textbf{\textit{-0.049}} & \textbf{\textit{0.114}} & \textbf{\textit{-0.157}} & \textbf{\textit{1.363}} & -0.036 & \textbf{\textit{1.314}} \\
Property \& Facilities & 1,901 & \textbf{-0.021} & \textit{0.021} & -0.015 & \textbf{0.244} & -0.198 & 0.438 \\
Sales \& Marketing & 5,416 & \textbf{\textit{0.058}} & \textbf{\textit{0.125}} & \textbf{\textit{0.059}} & \textbf{\textit{0.877}} & \textbf{\textit{0.176}} & \textbf{0.447} \\
Specialized Services & 1,171 & \textbf{0.025} & \textbf{\textit{-0.122}} & \textbf{\textit{0.050}} & \textbf{\textit{-0.879}} & \textbf{\textit{0.726}} & \textbf{\textit{-1.940}} \\
Strategy & 415 & \textbf{\textit{0.059}} & \textbf{\textit{0.272}} & -0.012 & \textbf{\textit{2.830}} & -0.379 & \textbf{\textit{3.872}} \\
\bottomrule
\end{tabular}

%% file: tables/t4a.tex
\begin{tabular}{lrrrrrrrrr}
\toprule
Month & N & \WRITINGCOMMUNICATION & \CODINGDATAANALYSIS & \TEXTBASEDANALYSIS & \KNOWLEDGEEXPERTISE & \SOFTWARETOOLGUIDANCE & \CREATIVETHINKINGIDEATION & \PERSONALNONWORK & \OTHER \\
\midrule
Sample Mean & 158,496 & 0.733 & 0.073 & 0.107 & 0.230 & 0.099 & 0.018 & 0.019 & 0.012 \\
2025-01 & 15,992 & \textbf{\textit{0.025}} & \textbf{\textit{-0.014}} & -0.002 & 0.003 & \textbf{\textit{-0.011}} & \textbf{\textit{0.004}} & \textbf{\textit{-0.003}} & -0.001 \\
2025-02 & 19,775 & \textbf{\textit{0.013}} & \textbf{\textit{-0.009}} & \textbf{\textit{0.007}} & \textbf{\textit{0.011}} & \textit{-0.004} & \textbf{\textit{0.006}} & \textbf{\textit{-0.004}} & -0.000 \\
2025-03 & 21,641 & -0.003 & 0.001 & \textit{0.004} & \textbf{\textit{0.020}} & -0.002 & \textbf{\textit{0.002}} & \textbf{\textit{-0.003}} & \textbf{\textit{0.002}} \\
2025-04 & 22,144 & \textbf{-0.007} & \textbf{\textit{0.007}} & 0.000 & \textbf{\textit{-0.011}} & \textbf{0.004} & -0.001 & 0.000 & 0.000 \\
2025-05 & 20,270 & -0.000 & \textbf{\textit{0.006}} & \textbf{\textit{0.016}} & \textit{0.005} & -0.001 & \textbf{\textit{-0.003}} & 0.001 & -0.001 \\
2025-06 & 21,553 & \textbf{\textit{0.013}} & -0.001 & \textbf{\textit{0.009}} & \textbf{\textit{-0.015}} & \textbf{\textit{-0.007}} & \textbf{\textit{-0.004}} & -0.001 & -0.001 \\
2025-07 & 17,381 & \textbf{\textit{-0.011}} & -0.000 & \textbf{\textit{-0.027}} & \textit{-0.006} & \textbf{0.005} & \textit{-0.002} & \textbf{\textit{0.004}} & \textbf{0.002} \\
2025-08 & 19,740 & \textbf{\textit{-0.026}} & \textbf{\textit{0.005}} & \textbf{\textit{-0.013}} & \textbf{-0.007} & \textbf{\textit{0.014}} & \textbf{-0.002} & \textbf{\textit{0.006}} & -0.001 \\
\bottomrule
\end{tabular}

%% file: tables/t4b.tex
\begin{tabular}{lrrrrrrrrr}
\toprule
Level & N & \WRITINGCOMMUNICATION & \CODINGDATAANALYSIS & \TEXTBASEDANALYSIS & \KNOWLEDGEEXPERTISE & \SOFTWARETOOLGUIDANCE & \CREATIVETHINKINGIDEATION & \PERSONALNONWORK & \OTHER \\
\midrule
Sample Mean & 131,736 & 0.756 & 0.052 & 0.104 & 0.240 & 0.091 & 0.020 & 0.021 & 0.012 \\
Staff & 49,517 & \textbf{\textit{0.022}} & \textbf{\textit{-0.005}} & \textbf{\textit{-0.007}} & \textbf{\textit{-0.053}} & \textbf{\textit{-0.007}} & 0.000 & \textbf{\textit{0.009}} & \textbf{\textit{0.003}} \\
Manager & 33,723 & \textbf{\textit{-0.043}} & \textbf{\textit{0.037}} & 0.001 & \textbf{\textit{-0.027}} & \textbf{\textit{0.034}} & \textbf{\textit{-0.003}} & \textbf{\textit{-0.008}} & \textbf{\textit{-0.001}} \\
Above Manager & 48,496 & \textbf{\textit{0.007}} & \textbf{\textit{-0.020}} & \textbf{\textit{0.007}} & \textbf{\textit{0.073}} & \textbf{\textit{-0.016}} & \textbf{\textit{0.002}} & \textbf{\textit{-0.003}} & \textbf{\textit{-0.002}} \\
\bottomrule
\end{tabular}

%% file: tables/t4c.tex
\begin{tabular}{lrrrrrrrrr}
\toprule
\makecell[l]{Functional\\Area} & N & \WRITINGCOMMUNICATION & \CODINGDATAANALYSIS & \TEXTBASEDANALYSIS & \KNOWLEDGEEXPERTISE & \SOFTWARETOOLGUIDANCE & \CREATIVETHINKINGIDEATION & \PERSONALNONWORK & \OTHER \\
\midrule
Sample Mean & 131,736 & 0.756 & 0.052 & 0.104 & 0.240 & 0.091 & 0.020 & 0.021 & 0.012 \\
Accounting \& Finance & 8,759 & \textbf{\textit{-0.124}} & \textbf{\textit{0.021}} & \textbf{\textit{-0.032}} & \textbf{\textit{0.040}} & \textbf{\textit{0.059}} & \textbf{\textit{-0.009}} & -0.000 & \textit{-0.002} \\
Administrative Services & 14,172 & \textbf{\textit{0.050}} & \textbf{\textit{-0.029}} & \textbf{\textit{-0.018}} & \textbf{\textit{-0.058}} & \textbf{\textit{-0.020}} & \textbf{\textit{-0.003}} & \textbf{\textit{0.020}} & \textbf{\textit{0.004}} \\
Affiliate Operations & 1,628 & \textbf{\textit{0.083}} & -0.008 & \textbf{\textit{-0.032}} & \textbf{\textit{-0.095}} & \textbf{\textit{-0.029}} & \textbf{\textit{-0.009}} & \textbf{\textit{-0.011}} & -0.003 \\
Communications & 4,273 & \textbf{\textit{0.118}} & \textbf{\textit{-0.039}} & -0.004 & \textbf{\textit{-0.031}} & \textbf{\textit{-0.071}} & \textbf{\textit{0.009}} & \textbf{\textit{-0.010}} & \textbf{\textit{-0.006}} \\
Compliance & 2,991 & \textbf{\textit{0.051}} & \textbf{\textit{-0.036}} & \textbf{\textit{-0.031}} & \textbf{\textit{0.039}} & \textbf{\textit{-0.029}} & \textbf{\textit{-0.009}} & \textbf{-0.005} & -0.002 \\
Digital Innovation & 1,464 & \textbf{\textit{-0.114}} & \textbf{\textit{-0.021}} & \textbf{\textit{0.076}} & \textbf{\textit{0.103}} & \textbf{\textit{-0.046}} & \textbf{0.011} & \textbf{\textit{-0.010}} & 0.000 \\
Information Technology & 19,115 & \textbf{\textit{-0.182}} & \textbf{\textit{0.109}} & \textbf{\textit{-0.031}} & \textbf{\textit{-0.008}} & \textbf{\textit{0.112}} & \textbf{\textit{-0.008}} & \textbf{\textit{-0.004}} & \textbf{\textit{0.002}} \\
Internal Audit & 915 & 0.003 & \textbf{\textit{-0.037}} & \textbf{-0.021} & \textbf{\textit{0.167}} & -0.005 & \textbf{\textit{-0.012}} & \textbf{-0.009} & 0.007 \\
Internal Operations & 11,990 & \textbf{\textit{0.045}} & \textbf{\textit{-0.025}} & \textbf{\textit{-0.033}} & \textbf{\textit{-0.070}} & \textbf{\textit{-0.023}} & \textbf{\textit{-0.003}} & \textbf{\textit{0.027}} & \textbf{\textit{0.003}} \\
Other & 16,383 & \textbf{\textit{0.040}} & \textbf{\textit{-0.026}} & \textbf{\textit{0.015}} & \textbf{\textit{0.017}} & \textbf{\textit{-0.016}} & -0.001 & \textbf{\textit{-0.005}} & -0.001 \\
Project Management & 6,282 & \textbf{\textit{0.023}} & \textbf{\textit{-0.031}} & \textbf{\textit{0.054}} & \textbf{\textit{0.114}} & \textbf{\textit{-0.009}} & -0.002 & \textbf{\textit{-0.007}} & -0.001 \\
Property \& Facilities & 8,053 & \textbf{\textit{0.052}} & \textbf{\textit{-0.017}} & \textbf{\textit{-0.026}} & \textbf{\textit{-0.041}} & \textbf{\textit{-0.015}} & \textbf{\textit{0.008}} & \textbf{0.004} & -0.001 \\
Sales \& Marketing & 29,149 & \textbf{\textit{0.062}} & \textbf{\textit{-0.016}} & \textbf{\textit{0.031}} & \textbf{\textit{0.009}} & \textbf{\textit{-0.046}} & \textbf{\textit{0.010}} & \textbf{\textit{-0.009}} & \textbf{-0.001} \\
Specialized Services & 2,523 & \textbf{\textit{-0.086}} & \textbf{\textit{0.035}} & \textbf{\textit{-0.025}} & \textit{0.017} & \textbf{\textit{0.091}} & \textbf{\textit{-0.008}} & \textbf{\textit{-0.010}} & -0.002 \\
Strategy & 4,039 & \textbf{\textit{-0.028}} & \textbf{\textit{-0.016}} & \textbf{\textit{0.092}} & \textbf{\textit{0.091}} & \textbf{-0.009} & 0.002 & \textbf{\textit{-0.014}} & \textbf{\textit{-0.004}} \\
\bottomrule
\end{tabular}

%% file: tables/t6a.tex
\begin{tabular}{lrrrr}
\toprule
Group & N & \CLARITYZ & \ANYSTRATEGY & \USECASEDIVERSITY \\
\midrule
Sample Mean & 17,671 & 0.000 & 0.058 & 0.316 \\
2025-01 & 1,799 & \textbf{\textit{-0.078}} & -0.003 & \textbf{\textit{-0.023}} \\
2025-02 & 2,048 & -0.031 & \textit{0.007} & 0.005 \\
2025-03 & 2,191 & 0.017 & \textbf{0.008} & \textbf{\textit{0.012}} \\
2025-04 & 2,262 & -0.019 & \textit{0.006} & \textit{0.007} \\
2025-05 & 2,257 & 0.017 & \textbf{\textit{-0.011}} & 0.001 \\
2025-06 & 2,477 & \textbf{\textit{0.071}} & \textbf{\textit{-0.011}} & -0.006 \\
2025-07 & 2,267 & 0.004 & 0.002 & -0.007 \\
2025-08 & 2,370 & -0.006 & 0.002 & \textit{0.007} \\
\bottomrule
\end{tabular}

%% file: tables/t6b.tex
\begin{tabular}{lrrrr}
\toprule
Group & N & \CLARITYZ & \ANYSTRATEGY & \USECASEDIVERSITY \\
\midrule
Sample Mean & 14,449 & 0.002 & 0.061 & 0.322 \\
Staff & 5,206 & \textbf{\textit{-0.056}} & \textbf{\textit{-0.009}} & \textbf{\textit{-0.011}} \\
Manager & 3,444 & 0.023 & 0.002 & 0.002 \\
Above Manager & 5,799 & \textbf{\textit{0.036}} & \textbf{\textit{0.007}} & \textbf{\textit{0.009}} \\
\bottomrule
\end{tabular}

%% file: tables/t6c.tex
\begin{tabular}{lrrrr}
\toprule
Group & N & \CLARITYZ & \ANYSTRATEGY & \USECASEDIVERSITY \\
\midrule
Sample Mean & 14,449 & 0.002 & 0.061 & 0.322 \\
Accounting \& Finance & 1,192 & \textbf{\textit{-0.311}} & \textbf{-0.010} & \textbf{\textit{-0.019}} \\
Administrative Services & 1,407 & \textbf{\textit{-0.159}} & \textbf{\textit{-0.013}} & \textbf{-0.015} \\
Affiliate Operations & 207 & \textbf{-0.139} & -0.013 & \textbf{\textit{-0.050}} \\
Communications & 461 & \textbf{\textit{0.343}} & 0.007 & \textbf{\textit{-0.024}} \\
Compliance & 299 & -0.074 & 0.014 & -0.016 \\
Digital Innovation & 190 & \textbf{\textit{0.302}} & -0.002 & \textbf{\textit{0.061}} \\
Information Technology & 2,552 & \textbf{\textit{-0.127}} & \textbf{\textit{0.011}} & \textbf{\textit{0.026}} \\
Internal Audit & 100 & \textbf{-0.203} & -0.003 & 0.009 \\
Internal Operations & 1,309 & \textbf{\textit{-0.094}} & \textbf{\textit{-0.024}} & \textbf{\textit{-0.026}} \\
Other & 1,572 & \textbf{\textit{0.133}} & 0.007 & 0.004 \\
Project Management & 574 & \textbf{\textit{0.122}} & \textbf{0.019} & \textbf{\textit{0.038}} \\
Property \& Facilities & 872 & \textbf{\textit{-0.075}} & \textbf{\textit{-0.025}} & \textbf{-0.016} \\
Sales \& Marketing & 3,051 & \textbf{\textit{0.214}} & \textbf{\textit{0.009}} & -0.001 \\
Specialized Services & 369 & \textbf{\textit{-0.154}} & 0.003 & -0.011 \\
Strategy & 294 & \textbf{\textit{0.303}} & 0.010 & \textbf{\textit{0.056}} \\
\bottomrule
\end{tabular}

%% file: tables/t7a.tex
\begin{tabular}{lrrrrrrr}
\toprule
\multicolumn{2}{r}{} & \multicolumn{2}{r}{\CLARITYZ} & \multicolumn{2}{r}{\ANYSTRATEGY} & \multicolumn{2}{r}{\USECASEDIVERSITY} \\
Predictor & N & beta & p & beta & p & beta & p \\
\midrule
\multicolumn{8}{l}{\textit{Ambition}} \\
\quad \UMAVGFIRSTUSERCHARS & 17,671 & 0.000 & 0.000 & 0.000 & 0.000 & 0.000 & 0.000 \\
\addlinespace
\multicolumn{8}{l}{\textit{Persistence}} \\
\quad \LOGUMAVGUSERROUNDS & 17,671 & 0.279 & 0.000 & 0.021 & 0.000 & 0.054 & 0.000 \\
\quad \USERROUNDSGETWO & 17,671 & 0.323 & 0.000 & 0.024 & 0.000 & 0.080 & 0.000 \\
\quad \USERROUNDSGETHREE & 17,671 & 0.328 & 0.000 & 0.034 & 0.000 & 0.056 & 0.000 \\
\quad \USERROUNDSGEFIVE & 17,671 & 0.333 & 0.000 & 0.032 & 0.000 & 0.036 & 0.000 \\
\addlinespace
\multicolumn{8}{l}{\textit{Frequency}} \\
\quad \LOGUMNUMCONVERSATIONS & 17,671 & 0.134 & 0.000 & -0.008 & 0.000 & 0.109 & 0.000 \\
\quad \LOGGPTUSAGEDATES & 14,449 & 0.101 & 0.000 & -0.023 & 0.000 & 0.121 & 0.000 \\
\quad \GPTDAYSGEFIVE & 14,449 & 0.148 & 0.000 & -0.029 & 0.000 & 0.164 & 0.000 \\
\quad \GPTDAYSGETEN & 14,449 & 0.125 & 0.000 & -0.026 & 0.000 & 0.113 & 0.000 \\
\quad \GPTDAYSGEFIFTEEN & 14,449 & 0.108 & 0.002 & -0.026 & 0.000 & 0.071 & 0.000 \\
\addlinespace
\multicolumn{8}{l}{\textit{Flexibility}} \\
\quad \LOGCOPILOTUSAGEDATES & 14,449 & 0.148 & 0.000 & 0.014 & 0.000 & 0.038 & 0.000 \\
\quad \COPILOTDAYSGEONE & 14,449 & 0.283 & 0.000 & 0.029 & 0.000 & 0.046 & 0.000 \\
\quad \COPILOTDAYSGEFIVE & 14,449 & 0.229 & 0.000 & 0.021 & 0.000 & 0.069 & 0.000 \\
\quad \COPILOTDAYSGETEN & 14,449 & 0.241 & 0.000 & 0.022 & 0.000 & 0.067 & 0.000 \\
\quad \COPILOTDAYSGEFIFTEEN & 14,449 & 0.258 & 0.000 & 0.020 & 0.001 & 0.068 & 0.000 \\
\addlinespace
\multicolumn{8}{l}{\textit{Training}} \\
\quad \LOGTRAININGSCOMPLETEDMONTH & 13,091 & 0.084 & 0.000 & 0.054 & 0.000 & 0.037 & 0.000 \\
\quad \TRAININGSMONTHGEONE & 13,091 & 0.069 & 0.000 & 0.056 & 0.000 & 0.034 & 0.000 \\
\quad \TRAININGSMONTHGETHREE & 13,091 & 0.157 & 0.002 & 0.046 & 0.000 & 0.048 & 0.000 \\
\quad \TRAININGSMONTHGEFIVE & 13,091 & 0.250 & 0.003 & 0.061 & 0.001 & 0.039 & 0.022 \\
\quad \LOGTRAININGSCOMPLETEDSTARTMONTH & 13,091 & 0.159 & 0.000 & 0.018 & 0.000 & 0.028 & 0.000 \\
\quad \TRAININGSSTARTGEONE & 13,091 & 0.130 & 0.000 & 0.016 & 0.003 & 0.029 & 0.000 \\
\quad \TRAININGSSTARTGETHREE & 13,091 & 0.194 & 0.000 & 0.017 & 0.002 & 0.032 & 0.000 \\
\quad \TRAININGSSTARTGEFIVE & 13,091 & 0.231 & 0.000 & 0.021 & 0.002 & 0.039 & 0.000 \\
\quad \TRAININGSSTARTGETEN & 13,091 & 0.225 & 0.007 & 0.022 & 0.132 & 0.032 & 0.029 \\
\bottomrule
\end{tabular}

%% file: tables/t7b.tex
\begin{tabular}{lrrrrrrr}
\toprule
\multicolumn{2}{r}{} & \multicolumn{2}{r}{\CLARITYZ} & \multicolumn{2}{r}{\ANYSTRATEGY} & \multicolumn{2}{r}{\USECASEDIVERSITY} \\
Predictor & N & beta & p & beta & p & beta & p \\
\midrule
\multicolumn{8}{l}{\textit{Ambition}} \\
\quad \UMAVGFIRSTUSERCHARS & 17,671 & 0.000 & 0.000 & 0.000 & 0.012 & 0.000 & 0.000 \\
\addlinespace
\multicolumn{8}{l}{\textit{Persistence}} \\
\quad \LOGUMAVGUSERROUNDS & 17,671 & 0.289 & 0.000 & 0.032 & 0.000 & 0.058 & 0.000 \\
\quad \USERROUNDSGETWO & 17,671 & 0.244 & 0.000 & 0.028 & 0.000 & 0.053 & 0.000 \\
\quad \USERROUNDSGETHREE & 17,671 & 0.250 & 0.000 & 0.034 & 0.000 & 0.043 & 0.000 \\
\quad \USERROUNDSGEFIVE & 17,671 & 0.256 & 0.000 & 0.028 & 0.000 & 0.032 & 0.000 \\
\addlinespace
\multicolumn{8}{l}{\textit{Frequency}} \\
\quad \LOGUMNUMCONVERSATIONS & 17,671 & 0.108 & 0.000 & 0.010 & 0.000 & 0.145 & 0.000 \\
\quad \LOGGPTUSAGEDATES & 14,449 & 0.066 & 0.000 & -0.002 & 0.522 & 0.159 & 0.000 \\
\quad \GPTDAYSGEFIVE & 14,449 & 0.074 & 0.000 & -0.001 & 0.877 & 0.150 & 0.000 \\
\quad \GPTDAYSGETEN & 14,449 & 0.038 & 0.044 & -0.000 & 0.931 & 0.080 & 0.000 \\
\quad \GPTDAYSGEFIFTEEN & 14,449 & 0.055 & 0.031 & -0.005 & 0.160 & 0.043 & 0.000 \\
\addlinespace
\multicolumn{8}{l}{\textit{Flexibility}} \\
\quad \LOGCOPILOTUSAGEDATES & 14,449 & 0.005 & 0.743 & 0.004 & 0.159 & 0.035 & 0.000 \\
\quad \COPILOTDAYSGEONE & 14,449 & 0.046 & 0.101 & 0.012 & 0.028 & 0.024 & 0.003 \\
\quad \COPILOTDAYSGEFIVE & 14,449 & -0.011 & 0.621 & 0.003 & 0.516 & 0.034 & 0.000 \\
\quad \COPILOTDAYSGETEN & 14,449 & -0.005 & 0.832 & -0.002 & 0.763 & 0.037 & 0.000 \\
\quad \COPILOTDAYSGEFIFTEEN & 14,449 & 0.028 & 0.306 & -0.002 & 0.774 & 0.034 & 0.000 \\
\addlinespace
\multicolumn{8}{l}{\textit{Training}} \\
\quad \LOGTRAININGSCOMPLETEDMONTH & 13,091 & 0.071 & 0.000 & 0.040 & 0.000 & 0.036 & 0.000 \\
\quad \TRAININGSMONTHGEONE & 13,091 & 0.062 & 0.000 & 0.038 & 0.000 & 0.033 & 0.000 \\
\quad \TRAININGSMONTHGETHREE & 13,091 & 0.131 & 0.004 & 0.037 & 0.000 & 0.037 & 0.000 \\
\quad \TRAININGSMONTHGEFIVE & 13,091 & 0.103 & 0.226 & 0.039 & 0.027 & 0.041 & 0.017 \\
\quad \LOGTRAININGSCOMPLETEDSTARTMONTH & 13,091 & -0.033 & 0.179 & -0.025 & 0.000 & -0.008 & 0.183 \\
\quad \TRAININGSSTARTGEONE & 13,091 & -0.024 & 0.359 & -0.006 & 0.320 & 0.006 & 0.336 \\
\quad \TRAININGSSTARTGETHREE & 13,091 & 0.015 & 0.577 & -0.025 & 0.000 & -0.016 & 0.022 \\
\quad \TRAININGSSTARTGEFIVE & 13,091 & -0.070 & 0.046 & -0.021 & 0.004 & -0.013 & 0.148 \\
\quad \TRAININGSSTARTGETEN & 13,091 & -0.124 & 0.087 & -0.029 & 0.034 & -0.028 & 0.067 \\
\bottomrule
\end{tabular}

%% file: tables/MetaPrompts.tex
\subsection*{Prompt One: Use Cases}
\begin{Verbatim}[breaklines=true,breakanywhere=true,fontsize=\small]

## LLM Conversation Analysis Prompt

You are an LLM conversation analysis expert. You will analyze a multi-turn conversation between a KPMG employee and an LLM to evaluate its use cases.

Your analysis will classify the use case(s) of the conversation.

## USE CASE CLASSIFICATION

Classify the conversation into one or more subcategories. A single conversation often serves multiple purposes - capture all that apply.

## CRITICAL CLASSIFICATION PRINCIPLES
1. **Focus on the ACTION requested**, not just the subject matter discussed
  - Example - Describing data from documents that mention software → Document Understanding (3-1), not Software guidance
2. **Read the ENTIRE conversation**, not just initial prompts - user intent often clarifies through the dialogue
3. **Use multiple categories**, when applicable, most conversations warrant more than one.

## EVIDENCE POLICY FOR USING LLM RESPONSES
Classification decisions must rely only on the user prompts (the Initial User Prompt and any Follow Up User Prompts). You may look at LLM responses solely to interpret what the user is clarifying in subsequent follow-ups

### MAIN CATEGORIES AND SUBCATEGORIES:

**1: Writing and Communication**
- 1-1: Generate new content
  - **Includes**: Writing from scratch, creating based on requirements, expanding bullet points into full documents, creating summaries intended as deliverables
  - **Excludes**: Editing text that the USER provided (that's 1-2), summaries purely for comprehension (that's 3-1 only)
  - **Expansion rule**: If output is >2x longer than input = 1-1
 
- 1-2: Edit/improve existing content
  - **What counts as "existing content"**:
  - Text explicitly provided by the user (pasted, uploaded, or quoted)
  - Text the user claims to have written themselves
  - Full paragraphs or complete documents
  - Does NOT include: Brief examples, fragments, or bullet points that need expansion
  - **Includes**: Grammar fixes, clarity improvements, restructuring of provided text
  - **Excludes**: Creating new content based on descriptions or outlines
  - **Note**: Editing user-provided content about specialized topics (e.g., legal, accounting) is ONLY 1-2, and not also a knowledge category (4-x) unless the user requests development or expansion of the technical writing with domain knowledge, or requests an accuracy check.
**When to use BOTH 1-1 and 1-2**:
  - User provides some content AND requests significant new additions
  - Examples:
    - "Improve this text and add a conclusion" → both 1-1 and 1-2
    - "Enhance this section and add supporting examples" → both 1-1 and 1-2
    - If one prompt generates and another edits, then it's 1-1 and 1-2.

- 1-3: Language translation: Converting text between languages

- 1-other: Other writing tasks not listed above

**2: Coding and Data Analysis**
- 2-1: Generate or debug programming code
  - **Includes**: ANY executable code generation (Python, R, SQL, VBA, JavaScript, etc.) including VBA code for Excel
  - **Includes**: Code debugging, modification, optimization
  - **Excludes**: Discussing code concepts without generating code. Also, Excel formulas are not code in this taxonomy and should be classified as 5-1.

- 2-2: Analyze user-provided data
  - **REQUIRES**: Actual data provided (Excel, CSV, tables) AND analysis performed
  - **Includes**: Statistical analysis, data interpretation, insights from provided data
  - **Excludes**: Just discussing analysis methodology, manual calculations without data
  - **Excludes**: Analysis of user-provided writing/prose (that's 3-1)
  - **Data vs. Text Distinction**: If a table is text-only and fits on one page, treat as writing (1-x categories). Otherwise, treat as data (2-x categories)

- 2-3: Data cleaning, restructuring, standardization
  - **REQUIRES**: Actual data provided (Excel, CSV, tables) AND manipulation tasks performed
  - **Includes**: Removing duplicates, reformatting, normalizing data
  - **Key test**: Is the LLM actively transforming data? Not just discussing how to do it

- 2-other: Other coding/data tasks

**3: Text-Based Analysis**
- 3-1: Document Understanding
  - **REQUIRES**: Document specified AND requests made related to the content
  - **Note:** Document can be *specified* by being uploaded, pasted, or requested via search
  - **Includes**: "Summarize the following", "what does this say about X", "read this and prepare to answer questions", analysis of user-provided writing or prose
  - **Excludes**: General questions without documents, uploading text as an example (e.g., write in the style demonstrated by this document)
  - **Data (2-2) vs Text (3-1)**: Where data is structured, tabular, or otherwise aggregable, text is unique narrative or prose. When an uploaded file contains both, code based on the actions requested. When the same request involves text-based and data analysis, code as both 2-2 and 3-1.

3-other: Other text-based analysis requests

**4: Knowledge and Expertise**
- 4-1: Accounting standards and guidance
  - **Includes**: GAAP, IFRS, audit procedures, financial reporting
  - **Excludes**: General business operations, non-financial audits
  - **Note**: Writing a new memo to explain an accounting standard would be both (1-1 and 4-1). User-provided accounting content for editing = 1-2 only.

- 4-2: Legal standards, Regulations, Compliance, and related concepts
  - **Includes**: Legal principles, interpretations, SEC guidance, industry-specific regulations, compliance requirements
  - **Note**: Uploading an SEC Comment Letter and requesting a summary would be both document understanding (3-1) and this category (4-2).

- 4-3: Business operations
  - **Includes**: How industry, products, or workflows function
  - **Excludes**: Specific aspects other than operations (e.g., people management or HR) → that's 4-4

- 4-4: Business Management
  - **Includes**: People management techniques, HR, Marketing, financing, incentive design
  - **Excludes**: Business operations (e.g., revenue and collection cycle)

- 4-other: knowledge queries outside 4-1 - 4-4.

**5: Software and Tool Guidance**
**CRITICAL FOR ALL 5-X CATEGORIES**: User must be seeking guidance on HOW TO USE the software, not just mentioning it. Questions about creating content (e.g., "How do I create an effective presentation in PowerPoint?") focus on content creation (1-x categories), not software usage.

- 5-1: Microsoft Excel
  - **Includes**: Excel Formula help, Excel feature guidance, Excel-specific troubleshooting
  - **Key test**: Is the question about Excel's functionality? Not just Excel-related
  - **Excludes**: Discussion of formulas, code, data that are not specifically about Excel

- 5-2: Alteryx
  - **Watch for**: Brief mentions can be overlooked - scan entire conversation
  - **Includes**: Workflow guidance, Alteryx-specific formulas

- 5-3: Microsoft PowerPoint
  - **Focus on**: PowerPoint SOFTWARE guidance, not presentation content
  - **Common trap**: Creating or Revising original presentation content → 1-1 or 1-2, not 5-3
  - **Excludes**: Power BI (different product)

- 5-4: Microsoft Word
  - **Specific to**: Word software features and functionality
  - **Excludes**: General document writing (→ 1-1)

- 5-5: Microsoft Outlook
  - **Includes**: Email, calendar, meeting features IN OUTLOOK
  - **Excludes**: email/calendar content without discussing Outlook features

- 5-6: AI/LLM tools
  - **STRICT CRITERION**: Discussion must be *about* the AI/LLM tool itself.
  - **Includes**: How to use AI tools, comparing AI tools, AI tool best practices, meta-conversations about AI usage
  - **Excludes**: Using an LLM to accomplish tasks (classify by the task instead)
  - **Example**: Which prompting techniques work best in Claude-3 for extracting structured data from PDFs? Compare with GPT-4o.

- 5-other: guidance on other software (Google Sheets, Tableau, etc.)

**6. Creative Thinking**
- 6-1: Brainstorming and ideation
  - **REQUIRES**: Generating NEW ideas, not listing existing ones
  - **Key words**: "ideas for", "possibilities", "creative solutions", "what if", propose other titles

- 6-other: other creative thinking requests

**7: Non-work Related**
- 7-1: Personal tasks
  - **Clear indicators**: Grocery lists, personal travel, home projects
  - **Note**: Even vague initial prompts often clarify as personal through conversation
  - **Example**: Revising a resume or anticipating interview questions for a job interview rather than bidding for a project.

- 7-other: Other non-work activities

**8: Other**
- 8-1: Testing LLM capabilities
  - **Intent**: Explicitly testing what the LLM can do

- 8-2: Unclear or ambiguous use case
  - **Use when**: Initial prompt lacks detail AND conversation doesn't clarify

- 8-other: anything that truly fits nowhere else

**### USING "OTHER" SUBCATEGORIES:**
- Use specific "other" subcategories (e.g., 1-other, 2-other) when:
 - The conversation clearly belongs to the main category
 - But doesn't fit any existing subcategory
 - Example: A writing task that isn't generating, editing, or translating
- Use 8-other only when the conversation doesn't fit ANY main category
- Always attempt to classify into existing subcategories before using "other" options

### DATA VS TEXT DISTINCTION:
1. **Read the ENTIRE conversation** before classifying - intent often clarifies
2. **Focus on actions and outcomes**, not just topics discussed
3. **Use multiple subcategories**, when appropriate

### EXEMPLARS

1. **Domain‑knowledge tagging rule**
  - **ADD a 4-x tag when**:
    - User explicitly requests domain expertise to be incorporated ("explain the accounting treatment for...")
    - User asks for accuracy verification of technical content
    - User requests application of standards/regulations to a scenario
    - The PRIMARY value comes from domain knowledge, not just writing ability
  - **DO NOT add a 4-x tag when**:
    - User provides the technical content and only requests formatting/editing
    - Domain knowledge is incidental to the writing task

2. **Business operations + management** *(double‑coding 4‑3 + 4‑4)*
  * Prompt: “Describe how a regional retail bank generates revenue across its core products (deposits, consumer lending, wealth management). Then outline a performance‑measurement and incentive‑compensation plan for branch managers that aligns with those revenue streams.”
  * Tags: **4‑3 + 4‑4**

3. **Example of the importance of reading the entire conversation and using multiple categories**:
  - If a user submits one prompt requesting a summary of user-provided text, that would be 3-1 (document understanding) as the intended use seems to be personal learning. By contrast, following the requested summary with requests for revisions provides evidence that the user's first prompt also wanted help with Writing and Communication; therefore, second prompt reveals that the original prompt is both document understanding (3-1), and generate new content (1-1). Then, the revision request in the second prompt is also edit/improve content (1-2). Note: This example demonstrates how later messages within the same conversation can reveal the true intent of initial prompts, highlighting why the entire conversation must be read before classification.

4. **Clear 1-1, 1-2, and 3-1 distinction**
 - "Create a one-page summary of this document for the board" → 1-1 and 3-1
 - "Help me understand this report" → 3-1 only
 - "Draft a summary based on this report" → 1-1 and 3-1
 - "Read the attached document and use it to revise the following message" → 1-2 and 3-1

## INPUT
You were or will be given a labeled conversation transcript with the following turn types (in order as applicable):
  1. System Prompt
  2. Initial User Prompt
  3. Initial LLM Response
  4. Follow Up User Prompt (zero or more)
  5. Follow Up LLM Response (interleaved with follow ups)
  6. Final LLM Response (if present)
Prompts and responses alternate chronologically. Do not use LLM Responses for classifications except to provide context for follow-up user prompts.

## OUTPUT FORMAT

Return your analysis as a single JSON object:

{
 "use_cases": ["1-1", "1-2", "4-4", "8-other"]
}

Notes:
- Replace the array values with every sub-category that applies to the conversation you just analyzed.

CRITICAL FORMATTING INSTRUCTIONS:
- Return ONLY the raw JSON object
- Do NOT wrap the JSON in markdown code blocks
- Do NOT include any markdown in your response
- DO NOT DO NOT DO NOT include \n before entries in the JSON
- Do NOT include any text before or after the JSON
- Start your response with { and end with }
- Ensure the JSON is valid and properly formatted

IMPORTANT: Return ONLY the JSON object. No other text, explanation, or formatting.
\end{Verbatim}

\bigskip\hrule\bigskip

\subsection*{Prompt Two: Strategies}
\begin{Verbatim}[breaklines=true,breakanywhere=true,fontsize=\small]
## LLM Conversation Analysis Prompt

You are an LLM conversation analysis expert. You will analyze a multi-turn conversation between a KPMG employee and an LLM to evaluate its prompting strategies.

### PROMPTING STRATEGIES:

Identify which of these prompting strategies the user employed, capture all that apply:
1. **Role-playing** (CODE: "role_play"): User explicitly assigns the LLM a specific identity, persona, or professional role
 - **Unique marker:** Direct identity assignment using "You are", "Act as", "Pretend to be", "Take the role of"
 - **Key distinction:** The LLM must be told to BE someone/something, not just write in a style

2. **Few-shot** (CODE: "few_shot"): User provides concrete examples of desired input-output pairs
 - **Unique marker:** Actual examples showing "if input is X, output should be Y"
 - **Key distinction:** Must show complete examples, not just templates or formats

3. **Chain-of-thought** (CODE: "chain_thought"): User requests to see the reasoning process or structured thinking
 - **Unique marker:** Requests for thinking/reasoning visibility: "show your work", "explain your reasoning", "think step-by-step", "walk through your logic", "break this down", "analyze step-by-step", "walk me through", "explain your analysis"
 - **Key distinction:** Focus is on exposing the cognitive process or structured analysis, not just organizing information

4. **Self-verification** (CODE: "self_verify"): User instructs LLM to check or validate its own output
 - **Unique marker:** Commands to self-review: "double-check your answer", "verify this is correct", "review for errors"
 - **Key distinction:** The LLM must be told to evaluate its own work, not just be careful

5. **Interactive refinement** (CODE: "interactive_refine"): User requests collaborative back-and-forth to refine the task
 - **Unique marker:** "help me think through this", "ask me questions", "let's figure this out together", "guide me through", "interview me about", "help me clarify my thoughts"
 - **Key distinction:** User seeks a collaborative dialogue rather than a response

### CLASSIFICATION RULES:
- Analyze the ENTIRE conversation, not just initial prompts
- A prompt can have MULTIPLE strategies, capture all that apply
- Only code strategies that are EXPLICITLY requested
- If none apply, mark as "none"
- Each strategy must have clear, unambiguous linguistic markers

### INTENSITY SCORING:
For each identified strategy, assign intensity:
- **Light** (1): Brief or minimal use
- **Moderate** (2): Clear but not elaborate use 
- **Heavy** (3): Extensive or sophisticated use

Example: "role_play_2" = moderate role-playing

## INPUT
You were or will be given a labeled conversation transcript with the following turn types (in order as applicable):
  1. System Prompt
  2. Initial User Prompt
  3. Initial LLM Response
  4. Follow Up User Prompt (zero or more)
  5. Follow Up LLM Response (interleaved with follow ups)
  6. Final LLM Response (if present)
Prompts and responses alternate chronologically. Do not use LLM Responses for classifications except to provide context for follow-up user prompts.

## OUTPUT FORMAT

Return your complete analysis as a single JSON object:

{
 "strategies": ["role_play_1", "chain_thought_2", "interactive_refine_3", "none"]
}

CRITICAL FORMATTING INSTRUCTIONS:
- Return ONLY the raw JSON object
- Do NOT wrap the JSON in markdown code blocks
- Do NOT include any markdown in your response
- DO NOT DO NOT DO NOT include \n before entries in the JSON
- Do NOT include any text before or after the JSON
- Start your response with { and end with }
- Ensure the JSON is valid and properly formatted

IMPORTANT: Return ONLY the JSON object. No other text, explanation, or formatting.
\end{Verbatim}

\bigskip\hrule\bigskip

\subsection*{Prompt Three: Other Factors}
\begin{Verbatim}[breaklines=true,breakanywhere=true,fontsize=\small]
## LLM Conversation Analysis Prompt

## Instructions

You are an LLM conversation analysis expert. You will analyze a multi-turn conversation between a user and an LLM to evaluate various traits.

### CRITICAL CLASSIFICATION PRINCIPLES

1. **Look at the ENTIRE conversation**, not just initial prompts — user intent often clarifies through the dialogue
2. **Effective clarification through dialogue should not be penalized** — rate the overall journey, not just initial clarity
3. **Focus on the user's instructional language** — Analyze how users frame requests, not the content they submit for work. Exclude from analysis: quoted text, documents being revised, or any material the user wants rewritten/translated/fixed. An exception to this principle applies to item 2 below.

### EVIDENCE POLICY FOR USING LLM RESPONSES

Base ratings only on the user prompts (the Initial User Prompt and any Follow Up User Prompts). You may look at LLM responses solely to interpret what the user is clarifying in subsequent follow-ups; do not use LLM response content as evidence for any rating or indicator.

### QUALITATIVE RATINGS (1–5 Scale)

1. **Initial Language Complexity**: Rate linguistic complexity of the FIRST substantive user prompt (based on wording and syntax, not task structure or domain expertise)
 - 1 = Very simple (common vocabulary, very short sentences, minimal modifiers)
 - 2 = Simple (basic vocabulary, mostly single-clause sentences, limited qualifiers)
 - 3 = Moderate (some technical terms or qualifiers, multi-clause sentences, standard punctuation)
 - 4 = Complex (dense phrasing, advanced vocabulary/jargon, nested clauses, frequent hedging)
 - 5 = Very complex (highly technical or jargon-heavy, convoluted syntax with multiple embedded clauses)
 - Note: Skip standard greetings; assess the language form only—not the complexity of the task requested
 - Per Principle #3: Exclude work product from analysis (see classification principles)

2. **Initial Complexity**: Rate sophistication of the FIRST substantive user prompt (based on structural complexity, not domain expertise)
 - 1 = Trivial (simple question like "What is X?")
 - 2 = Simple (straightforward task like "Explain Y" or "List Z")
 - 3 = Moderate (multi-part request like "Compare X and Y" or "How does A affect B?")
 - 4 = Complex (sophisticated analysis like "Evaluate X considering Y constraints")
 - 5 = Very complex (multi-layered like "Analyze X considering Y constraints while optimizing for Z")
 - Note: Skip standard greetings; rate first substantive request
 - **Exception to critical classification principle #3**: If the work product's inherent complexity directly impacts task difficulty (e.g., debugging quantum computing code vs. fixing a typo), note this context but still rate based on how clearly the user articulates the task

3. **Prompt Structure Quality**: Organizational signals independent of prompting strategy
 - 1 = Unstructured text block
 - 2 = Minimal structure (single paragraph, few separators)
 - 3 = Some structure (bullets or numbered steps; minor inconsistencies)
 - 4 = Well structured (clear sections/headings, ordered steps, placeholders for variables)
 - 5 = Highly structured (sections such as context → task → constraints → output spec; clear variable slots)
 - Note: Do not evaluate strategy choices (no CoT/few-shot judgment here).
 - Per Principle #3: Exclude work product from analysis (see classification principles)

4. **Output Format Clarity**: Clarity of requested output format/specification
 - 1 = No output guidance
 - 2 = Vague guidance (e.g., "summarize")
 - 3 = Basic format hints (e.g., "bullet list", "markdown")
 - 4 = Clear format spec (sections, headings, approximate length, audience/tone)
 - 5 = Precise spec (schema/template provided; acceptance criteria for correctness)

5. **User Prompting Sophistication**: Strategy-agnostic assessment of prompting skill
 - 1 = Naive (single vague request)
 - 2 = Basic (simple request with one clarification)
 - 3 = Intermediate (multi-part request, some constraints, uses structure)
 - 4 = Advanced (clear scoping, robust constraints, well-organized)
 - 5 = Expert (templates/variables, acceptance criteria, anticipates failure modes)
 - Note: Do not score specific strategies (few-shot/CoT) here.
 - Per Principle #3: Exclude work product from analysis (see classification principles)

6. **Specificity Score**: Rate clarity and precision of user prompts across the entire conversation
 - 1 = Very vague (unclear objectives, ambiguous requests)
 - 2 = Somewhat vague (general direction but lacking details)
 - 3 = Moderately specific (clear intent, some details missing)
 - 4 = Mostly specific (well-defined, minor ambiguities)
 - 5 = Very specific (precise, unambiguous, complete context)
 - Note: If user starts vague but clarifies well through dialogue, rate the overall journey positively rather than negatively rating the initial vagueness

### BINARY INDICATORS

7. **Document Uploaded**: Did the user upload any files/attachments?
 - Look for: File references, "uploaded", "the attached" or other language referring to an attachment
 - Exclude: reference to attachment is in quoted portion of text, rather than in the actual request (as described by classification principle #3).

8. **Constraints Present**: Did the initial substantive prompt include explicit constraints (length, schema, tone, audience, timeframe)?
 - true = includes at least one actionable constraint, i.e., a requirement that is measurable or structurally binding.
 - false = the prompt offers no constraints or has only generic style cues or vague format hints such as “Be concise,” “shorten this,” “provide a simple x,” “use bullet points,” “brief overview,” “high level only,” “keep it short,” “make it readable,” etc.

9. **Structured Output Requested**: Did the user explicitly request structured data output?
 - true = User explicitly asked for JSON, YAML, CSV, or Markdown tables with specific fields/schema
 - false = No explicit request for structured data output
 - Exclude: General code requests, unstructured lists, vague formatting requests, or keep response under x number of words.

10. **Acceptance Criteria Provided**: Did the user define success tests (e.g., must include X fields, match Y style, pass Z checks)?

11. **User Satisfied**: Did the user express explicit positive feedback?
 - Look for: "thanks", "perfect", "exactly what I needed", "great", "helpful"
 - Exclude: Neutral acknowledgments like "okay", "I see".
 - Key test: Positive feedback should be expected in the first sentence of a "Follow Up User Prompt", not later in that prompt, and not in the "Initial User Prompt"

12. **Frequent Typos**: Does the user make frequent typing errors?
 - true = 3+ typos across the conversation OR 2+ typos in a single message
 - false = Fewer typos than the threshold
 - Count: Obvious misspellings, transposed letters, missing/extra characters
 - Exclude: Intentional abbreviations (ur, thx), domain-specific terms, proper nouns, grammar issues

13. **All Lowercase**: Does the user write entirely in lowercase?
 - true = 100% of user messages are entirely lowercase (excluding exemptions below)
 - false = User uses any capitalization in their messages
 - Exclude from analysis: URLs, file paths, email addresses, code snippets, single-word responses, provided content as described by classification principle #3.

14. **Pleasantries Used**: Does the user include polite social expressions?
 - true = User includes at least one pleasantry in their own words
 - false = No pleasantries used
 - Look for: Greetings ("hello", "hi", "good morning"), closings ("thanks", "thank you", "appreciate it"), politeness markers ("please", "could you", "would you mind"), social niceties ("hope you're well", "how are you")
 - Exclude: Pleasantries within provided content as described by classification principle #3.

### ANALYSIS GUIDELINES
- Analyze the ENTIRE conversation (typically 2–20 turns)
- Consider the overall journey, not just initial state
- Be consistent in ratings across conversations

### FALLBACK BEHAVIOR

If any field is indeterminable, set a sensible default (e.g., false) rather than inventing data.

## Input
You were or will be given a labeled conversation transcript with the following turn types (in order as applicable):
1. **System Prompt**
2. **Initial User Prompt**
3. **Initial LLM Response**
4. **Follow Up User Prompt** (zero or more)
5. **Follow Up LLM Response** (interleaved with follow ups)
6. **Final LLM Response** (if present)
Prompts and responses alternate chronologically. Base ratings only on the user prompts; use LLM responses only for context of follow-ups.

## CRITICAL FORMATTING INSTRUCTIONS
- Return ONLY the raw JSON object
- Do NOT wrap the JSON in markdown code blocks
- Do NOT include any markdown in your response
- DO NOT include \n before entries in the JSON
- Do NOT include any text before or after the JSON
- Start your response with { and end with }
- Ensure the JSON is valid and properly formatted

## REQUIRED JSON SCHEMA

Provide the output as a single JSON object with exactly these keys and value types (listed in the same order as the rubric above):

{
 "initial_language_complexity": <integer 1-5>,
 "initial_complexity": <integer 1-5>,
 "prompt_structure_quality": <integer 1-5>,
 "output_format_clarity": <integer 1-5>,
 "user_prompting_sophistication": <integer 1-5>,
 "specificity_score": <integer 1-5>,
 "document_uploaded": <boolean>,
 "constraints_present": <boolean>,
 "structured_output_requested": <boolean>,
 "acceptance_criteria_provided": <boolean>,
 "user_satisfied": <boolean>,
 "frequent_typos": <boolean>,
 "all_lowercase": <boolean>,
 "pleasantries_used": <boolean>
}
\end{Verbatim}